\documentclass[sigconf, nonacm]{acmart}
\usepackage{balance}

\AtBeginDocument{%
  }

\begin{document}
\title{TactiDex: A Real-World Tactile-Guided Benchmark for Human-Like Dexterous Manipulation}

    \author{%
  Suting Ni$^{1,2}$ \quad
  Hanbing Zhang$^{1,2}$ \quad
  Zhenyu Wei$^{1}$ \quad
  Guo Chen$^{1}$ \quad
  Chixuan Zhang$^{1}$ \quad
  Ye Shi$^{1,2}$ \quad
  Jingya Wang$^{1,*}$\\[0.4em]
  \{nist2024, zhanghb2025, weizhy2024, chenguo2024, zhangchx12024,\\
  shiye, wangjingya\}@shanghaitech.edu.cn\\[0.5em]
  $^{1}$ Shanghaitech University \quad
  $^{2}$ InstAdapt\\[0.2em]
  }

\renewcommand{\shortauthors}{Ni et al.}

\begin{abstract}
  Tactile feedback is fundamental to Hand-Object Interaction (HOI), governing contact formation, force regulation, and stable manipulation, making it essential for achieving true human-like dexterous manipulation. Yet, current human-to-robot dexterous transfer pipelines primarily rely on kinematic trajectories, resulting in motion imitation without physically grounded interaction. To address this, we introduce TactiDex, a real-world tactile-guided benchmark specifically designed to move dexterous manipulation beyond kinematic mimicry toward contact-level human-likeness. TactiDex provides a comprehensive dataset that elegantly aligns whole-hand tactile signals with multi-granularity kinematic and object states, coupled with standardized evaluation metrics. Building upon this data paradigm, we propose a tactile-driven transfer framework that effectively translates human demonstrations into physically plausible robotic execution. 
  We introduce TactiSkill, a framework built upon a novel tri-component tactile reward that innovatively uses tactile signals as structured supervision. This reward unifies guidance, human-like alignment, and contact constraints into a single objective. Through comprehensive experiments on both single and bimanual tasks, we demonstrate that TactiSkill achieves superior performance in manipulation success and physical realism. This work lays a crucial foundation for advancing tactile-aware dexterous manipulation. Our project page at \url{https://tactidex.github.io/}.
\end{abstract}


\maketitle
{\let\thefootnote\relax\footnotetext{$^*$ Corresponding author: wangjingya@shanghaitech.edu.cn}}

\section{Introduction}
\label{sec:intro}
Hand-object interaction (HOI) lies at the core of embodied intelligence, enabling humans to manipulate tools, operate daily objects, and perform complex dexterous skills. A defining characteristic of human dexterity is the continuous regulation of contact through tactile feedback, which governs contact formation, force modulation, and manipulation stability. While vision provides spatial and geometric awareness \cite{wen2024foundationpose, fang2023anygrasp, wen2023bundlesdf}, tactile perception determines how physical interaction is established and maintained at the hand-object interface. Neuroscience and robotics research consistently show that successful manipulation relies heavily on force modulation rather than mere spatial positioning \cite{yin2025osmo, lin2023bi, qi2023hand, xia2022review}. Consequently, human dexterity is not solely a function of kinematic trajectories, but fundamentally grounded in contact-level physical interaction.

Transferring such human dexterity to robotic systems remains a longstanding challenge.
Recent advances in dexterous manipulation have explored human-to-robot skill transfer through imitation learning \cite{qin2022dexmv, wang2024dexcap, li2025maniptrans, cheng2024open, bharadhwaj2024roboagent}, motion retargeting  \cite{qin2023anyteleop, wang2025dexh2r, lakshmipathy2025kinematic, xin2026analyzing}, and reinforcement learning frameworks \cite{xu2023unidexgrasp, wang2022dexgraspnet}. These approaches typically align human hand kinematics with robotic embodiments, leveraging visual observations and joint-level motion supervision to reproduce task behaviors.
Despite impressive progress, this trajectory-centric paradigm inherently overlooks the tactile dynamics essential for regulating contact formation and force distribution. Consequently, robotic policies often successfully replicate superficial movement patterns, but deviate significantly in contact behavior, interaction stability, and force modulation.

A key factor underlying this gap is the absence of tactile-rich benchmarks and standardized evaluation protocols. Without strictly synchronized tactile-kinematic demonstrations, learning objectives are implicitly forced to prioritize motion matching over physically grounded interaction, leaving a substantial gap in achieving true human-like robotic dexterity.

To address this gap, we introduce \textbf{TactiDex}, a real-world tactile-rich HOI dataset for contact-aware human-to-robot dexterous transfer. TactiDex provides synchronized whole-hand tactile sensing, fine-grained hand kinematics, wrist 6D poses, and object 6D trajectories, forming a temporally aligned and contact-rich representation of human manipulation dynamics. Unlike prior datasets that focus primarily on visual observations or joint trajectories \cite{chao2021dexycb, fan2023arctic, yang2022oakink, zhan2024oakink2, liu2022hoi4d, liu2024taco}, TactiDex captures contact interactions between the hand and object through spatially distributed pressure measurements. The dataset spans diverse single-hand and bimanual manipulation tasks, covering varied object geometries and interaction patterns, thereby providing a broad testbed for studying tactile-aware transfer. Beyond data collection, we establish standardized contact-aware evaluation protocols that quantify contact fidelity, force alignment, and interaction stability. By integrating synchronized multi-granularity tactile and spatial states within a unified framework, TactiDex establishes a benchmark tailored to studying tactile-aware human-to-robot transfer.

Building upon this data paradigm, we propose \textbf{TactiSkill}, built on the \textbf{Tactile-Guided Tri-Component Transfer Framework} that leverages synchronized contact signals to guide robotic policy learning. Directly incorporating raw pressure measurements into reinforcement learning objectives, however, presents non-trivial challenges. Naïve formulations based solely on contact force matching often lead to unstable optimization, reward exploitation, or physically unrealistic behaviors in simulation \cite{liu2025vtdexmanip, kim2025tac2motion, zhang2023adaptive, field2025text2touch, bianchini2023simultaneous}. To address this, we introduce a tri-component tactile objective conceptualizing tactile signals as structured supervision, which integrates (1) contact guidance to encourage meaningful and timely contact formation, (2) human-like alignment to match force distribution patterns observed in human demonstrations, and (3) safety constraints to regularize excessive force and promote stable interaction. This structured formulation enables stable optimization while preserving contact-level fidelity, effectively moving dexterous transfer beyond kinematic mimicry toward physically grounded human-likeness.

In summary, our contributions are as follows:

(1) We introduce \textbf{TactiDex}, the first high-precision, long-horizon tactile-rich hand–object interaction dataset for dexterous bimanual manipulation, providing temporally synchronized whole-hand tactile signals, fine-grained hand kinematics, and object 6D trajectories, together with tactile-aware evaluation metrics for human-to-robot transfer.

(2) We propose \textbf{TactiSkill}, a tactile-guided human-to-robot transfer framework that leverages tactile sensing as a core supervisory signal for dexterous policy learning. To stabilize contact-rich interaction, we design a tri-component tactile reward framework that (i) encourages proactive contact formation, (ii) aligns tactile distributions with human demonstrations to achieve force-level human-likeness, and (iii) constrains excessive contact forces to ensure physically stable manipulation.

(3) Extensive experiments demonstrate that tactile-guided transfer significantly improves geometric robustness and physical fidelity across diverse manipulation tasks, and we further validate the effectiveness of TactiSkill through real-world dexterous hand deployment, showing that policies derived from our dataset can reproduce stable contact-rich manipulation on physical hardware.
\section{Related Work}
\label{sec:related}

\begin{table*}[tb]
    \caption{\textbf{Comparison with existing HOI datasets.} \textbf{High-Prec. Kinematics}: Global motion capture vs. vision-based estimation. \textbf{Tactile}: Whole-hand pressure maps. TactiDex uniquely unifies exact kinematics, physical tactile sensing, and text annotations for long-horizon bimanual manipulation.}
  \label{tab:datasets}
  \centering
  \resizebox{\textwidth}{!}{
  \begin{tabular}{@{}lccccccccc@{}}
    \toprule
    Dataset & Frame & Subj & Obj & Real-World & Bimanual & High-Prec. Kinematics & Tactile & Text Annot. & Long-horizon \\
    \midrule
    ObMan \cite{hasson2019learning}        & 154K & 20 & 3K     &              &              &  &              &              &              \\
    ContactPose \cite{wen2024foundationpose}  & 2.99M & 50 & 25 & $\checkmark$ & $\checkmark$ & $\checkmark$ &              &              &              \\
    HOI4D \cite{liu2022hoi4d}        & 3M & 9 & 1000 & $\checkmark$ &              &              &              &              &              \\
    DexYCB \cite{chao2021dexycb}       & 582K & 10 & 20 & $\checkmark$ &              & $\checkmark$ &              &              &              \\
    ARCTIC \cite{fan2023arctic}       & 2.1M & 19 & 11 & $\checkmark$ & $\checkmark$ & $\checkmark$ &              &              &              \\
    OakInk2 \cite{zhan2024oakink2}     & 4.01M & 9 & 75 & $\checkmark$ & $\checkmark$ & $\checkmark$ &              & $\checkmark$ & $\checkmark$  \\
    GRAB \cite{taheri2020grab}       & 1.62M & 10 & 51 & $\checkmark$ & $\checkmark$  & $\checkmark$ &              &   &              \\
    OPENTOUCH \cite{song2025opentouch}       & 0.54M & - & 800 & $\checkmark$ &   &  & $\checkmark$  &  $\checkmark$  &              \\
    \midrule
    \textbf{TactiDex (Ours)} & 5.1M & 10 & 49 & \textbf{$\checkmark$} & \textbf{$\checkmark$} & \textbf{$\checkmark$} & \textbf{$\checkmark$} & \textbf{$\checkmark$} & \textbf{$\checkmark$} \\
  \bottomrule
  \end{tabular}
  } 
\end{table*}

\subsection{Hand-Object Interaction Dataset}
\label{sec:related-hoi}
Numerous datasets facilitate hand-object interaction (HOI) modeling \cite{chao2021dexycb, fan2023arctic, yang2022oakink, zhan2024oakink2, liu2022hoi4d, liu2024taco, kwon2021h2o, tian2024gaze}. Motion-capture datasets \cite{fan2023arctic, zhan2024oakink2, chao2021dexycb, taheri2020grab} provide high-precision 3D trajectories for detailed manipulation analysis, while egocentric datasets \cite{liu2022hoi4d, hoque2025egodex} expand semantic diversity in open-world and instruction-conditioned settings. However, both paradigms primarily infer contact from geometric proximity or visual observation, leaving actual force modulation physically unmeasured. Synthetic benchmarks \cite{hasson2019learning} offer scalable contact signals but rely on simulated geometry, lacking real-world tactile fidelity. Most recently, OPENTOUCH \cite{song2025opentouch} introduced real-world whole-hand tactile sensing with text annotations; yet, it predominantly focuses on egocentric perception tasks (e.g., cross-modal retrieval and grasp classification) rather than dynamic manipulation control. Consequently, as summarized in Table~\ref{tab:datasets}, existing HOI datasets largely lack the tactile grounding necessary for physically interactive policy learning. To bridge this gap, we introduce \textbf{TactiDex}, a tactile-grounded HOI benchmark that synchronizes whole-hand tactile sensing with multi-granularity kinematic states and establishes standardized, contact-aware evaluation protocols, enabling physically grounded modeling of human dexterity beyond geometric imitation.

\subsection{Dexterous Manipulation \& Human-to-Robot Transfer}
\label{sec:related-dex}
Dexterous manipulation has witnessed substantial progress through model-based control, reinforcement learning, and large-scale policy learning \cite{xu2023unidexgrasp, wang2022dexgraspnet, arunachalam2023dexterous, li2023dexdeform, chen2024object}. Concurrently, learning from human demonstrations offers an intuitive pathway for transferring dexterity to robotic hands \cite{qin2022dexmv, wang2024dexcap}. Methods such as imitation learning and motion retargeting align trajectories under morphological constraints \cite{qin2023anyteleop, lakshmipathy2025kinematic}, while recent RL frameworks refine these transferred policies using reference motions and task-specific rewards \cite{dasari2023learning, li2025maniptrans, liu2024parameterized, makoviychuk2021isaac, chen2023bi}. Although these methods significantly improve trajectory reproduction, contact-level discrepancies frequently persist. Geometric alignment does not guarantee consistency in force modulation or pressure distribution, often resulting in unstable grasps, excessive force, or physically implausible interactions during delicate, contact-rich tasks. Therefore, while trajectory imitation is a strong starting point, it fails to fully resolve the physical interaction gap. In contrast, our approach introduces tactile-guided supervision that directly constrains contact formation and force distribution, shifting the human-to-robot transfer paradigm from purely geometric trajectory alignment to physically grounded interaction alignment.

\subsection{Tactile-Guided Policy Learning}
\label{sec:related-tactile}
Tactile sensing is increasingly recognized as a critical modality for robust manipulation. Early tactile-guided reinforcement learning (RL) applications predominantly utilize parallel-jaw grippers to solve foundational tasks like peg-in-hole insertions and terminal assemblies \cite{han2025zero, tang2025visual, li2025visuo, fu2023safe}. To improve generalizability, subsequent studies extensively explore visuo-tactile fusion and sim-to-real transfer paradigms to address the reality gap during unknown object manipulation \cite{ding2021sim, hansen2022visuotactile, su2024sim2real, huang20243d, ding2020sim, zhao2025touch, miller2025enhancing, liang2022visuo}. More recently, tactile RL has scaled to highly articulated multi-fingered dexterous hands \cite{kim2025tac2motion, hu2025dexterous}, accompanied by the development of multimodal pretraining benchmarks \cite{liu2025vtdexmanip} and attempts to leverage human tactile demonstrations for imitation learning \cite{yu2023mimictouch}. Despite this rapid progress, existing frameworks largely treat tactile signals merely as auxiliary observation vectors for unconstrained trial-and-error optimization, suffering fundamentally from the absence of synchronized, whole-hand hand-object interaction (HOI) physical datasets. To bridge this gap, our work introduces a true \textit{human-to-robot} tactile transfer paradigm by leveraging the real-world \textbf{TactiDex} dataset. Instead of learning contact heuristics from scratch, \textbf{TactiSkill} formulates high-fidelity human force distributions not as passive observations, but as active, structural physical priors. By directly supervising the force modulation of the robotic hand, our framework ensures the resulting policies transcend geometric mimicry to achieve physically grounded, human-like dexterity.
\section{TactiDex Dataset}
\label{sec:tactidex}
\begin{figure}[tb]
  \centering
  \includegraphics[width=\linewidth]{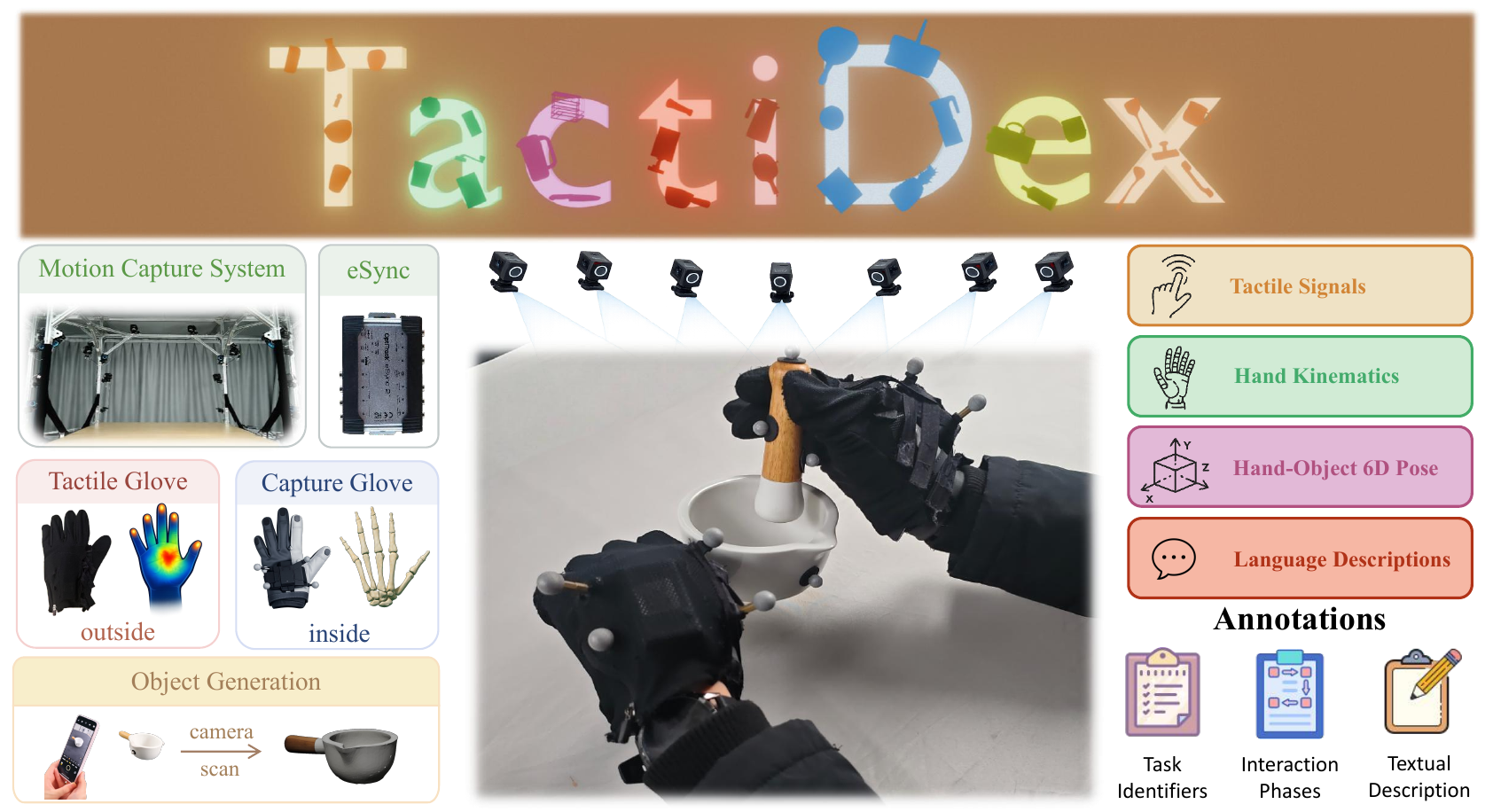}
  \caption{\textbf{Overview of the TactiDex Data Collection and Annotation Framework.} 
  (Left) Our hardware setup integrates a high-precision motion capture system with a novel dual-glove design---an inner capture glove for kinematics and an outer tactile glove for whole-hand pressure mapping, strictly synchronized via an eSync module. Manipulated objects are digitally reconstructed via 3D scanning. 
  (Center) A user performs complex, contact-rich interactions (e.g., bimanual grinding) within the capture volume. 
  (Right) The pipeline yields strictly synchronized multi-modal data streams, including tactile signals, hand kinematics, and hand-object 6D poses, enriched with comprehensive language descriptions and hierarchical task phase annotations.}
  \label{fig:TactiDex Dataset}
\end{figure}

\textbf{TactiDex}, as overviewed in Fig.~\ref{fig:TactiDex Dataset}, is a large-scale, real-world tactile-grounded dataset designed to capture the physical dynamics of human dexterous manipulation. It synchronizes whole-hand tactile sensing, articulated hand kinematics, wrist 6D poses, and object 6D trajectories within a unified motion-capture environment. The dataset comprises \textbf{49 calibrated objects} and \textbf{757 interaction sequences}, spanning from fundamental pick-and-place behaviors to complex functional operations and object-object bimanual interactions. By providing structured interaction annotations and task-level textual descriptions alongside multimodal sensory streams, TactiDex establishes a physically measurable and semantically grounded benchmark for contact-aware human-to-robot transfer.

\subsection{Hardware Setup}
\label{sec:hardware}
To develop the TactiDex dataset, we built a multimodal data collection system for tactile-grounded hand–object interaction that synchronously records hand contact information, hand kinematic states, and accurate object 6D poses in a global coordinate frame. The system consists of three hardware components: a full-hand tactile glove that captures pressure distributions, a hand-tracking glove that acquires articulated joint states, and an OptiTrack motion capture system that provides global 6-DoF poses for both the object and the hand.

\subsubsection{Tactile Measurement.}
\label{sec:tactile-glove}
The tactile gloves measure pressure via piezoresistive changes in flexible sensors. The tactile array comprises 162 sensing elements densely distributed across the fingertips and palm, with a force resolution of up to 0.01 N and a sampling rate of 17 Hz. We further redesigned the structure of tactile gloves and wearing configuration so that it can be worn externally over the motion capture glove. This modification enables stable acquisition of tactile signals from the whole hand without degrading the kinematic tracking quality of the mocap glove, while also minimizing interference with natural manipulation, improving reliability and consistency for long-duration data collection.

\subsubsection{Motion Capture.}
\label{sec:hand-tracking}
Hand kinematics are captured using a motion-tracking glove, where IMU measurements are fused with optical tracking data to estimate high-fidelity articulated joint states. Global 6-DoF poses for both the wrist and the objects are provided by an OptiTrack system equipped with 8 PX13W cameras operating at 120 fps within a 1.2 m $\times$ 1.8 m workspace. As both hand and object poses are processed within the same motion-capture environment, they are inherently aligned within a unified spatial coordinate system and temporal base.

\subsection{Data Collection and Processing}
\label{sec:data-collection}
\textbf{Synchronized Multimodal Capture.}
To ensure high-fidelity data acquisition, we established a strict collection protocol. Participants perform manipulation tasks within the calibrated multi-camera optical workspace. The hardware-synchronized pipelines ensure frame-level temporal alignment between the 120 Hz kinematic/pose streams and the tactile pressure data, enabling precise temporal modeling of contact formation and evolution.

\noindent \textbf{Object Modeling and Calibration.}
The 49 objects in our dataset were meticulously selected to cover diverse geometries, scales, and physical properties. Object meshes are reconstructed using a hybrid pipeline that integrates high-resolution mobile 3D scanning applications with SAM3D-based reconstruction methods \cite{chen2025sam}. To overcome the inherent feature-matching failures caused by reflective, transparent, or highly symmetric surfaces, we applied specialized non-reflective textured tapes and temporary fiducial markers during the scanning phase. This rigorous treatment ensured robust visual feature extraction, yielding high-fidelity triangular meshes regardless of challenging material properties. Following reconstruction, each object is dynamically calibrated within the OptiTrack system to ensure its 6D trajectory is spatially consistent with the hand's global coordinate frame.

\noindent \textbf{Task Coverage and Data Scope.}
We recorded 757 interaction sequences encompassing single-hand and bimanual manipulation. Task complexity ranges from basic physical interactions (e.g., pipkplace, inspect) to tool use and functional operations (e.g., click, cut, shake). Each sequence pairs the synchronized multimodal sensory data with structured annotations, including interaction-phase timestamps, task identifiers, and natural-language descriptions, providing a comprehensive foundation for multimodal policy learning.

\noindent \textbf{Optimization and Processing.}
Following OMOMO~\cite{li2023object}, we adopt the MANO model~\cite{romero2022embodied} for kinematic representation. We independently fit MANO parameters to the raw 3D motion-capture keypoints via L-BFGS optimization. To ensure anatomical and temporal consistency, we share shape parameters across all frames for each subject, and apply both an $\ell_2$ pose regularizer and a second-order temporal smoothness penalty to suppress high-frequency jitter.

While kinematic fitting captures accurate free-space trajectories, residual errors often cause physically implausible ``floating'' or penetration during object interaction. To resolve this, we introduce a two-stage \textit{tactile-constrained post-optimization}. 
In the first stage, we identify genuine contact intervals by gating geometric proximity with measured tactile pressure, effectively filtering out visually deceptive near-contacts. Within each interval, we select a stable reference grasp frame and optimize its spatial alignment using tactile-modulated per-finger weights: active fingers (exceeding the pressure threshold) are strongly attracted to the object surface, while inactive fingers are down-weighted to prevent phantom contacts. 
In the second stage, this refined grasp is propagated across the contact interval via temporal inverse kinematics, preserving original motions in non-contact regions. Finally, a collision-aware adjustment optimizes finger articulation against the object's Signed Distance Field (SDF) \cite{park2019deepsdf} to eliminate residual penetrations. This pipeline yields the final TactiDex annotations, ensuring that interactions are both kinematically precise and physically grounded by real-world tactile evidence.

\section{Methodology}
\label{sec:method}

\begin{figure*}[tb]
  \centering
  \includegraphics[width=\textwidth]{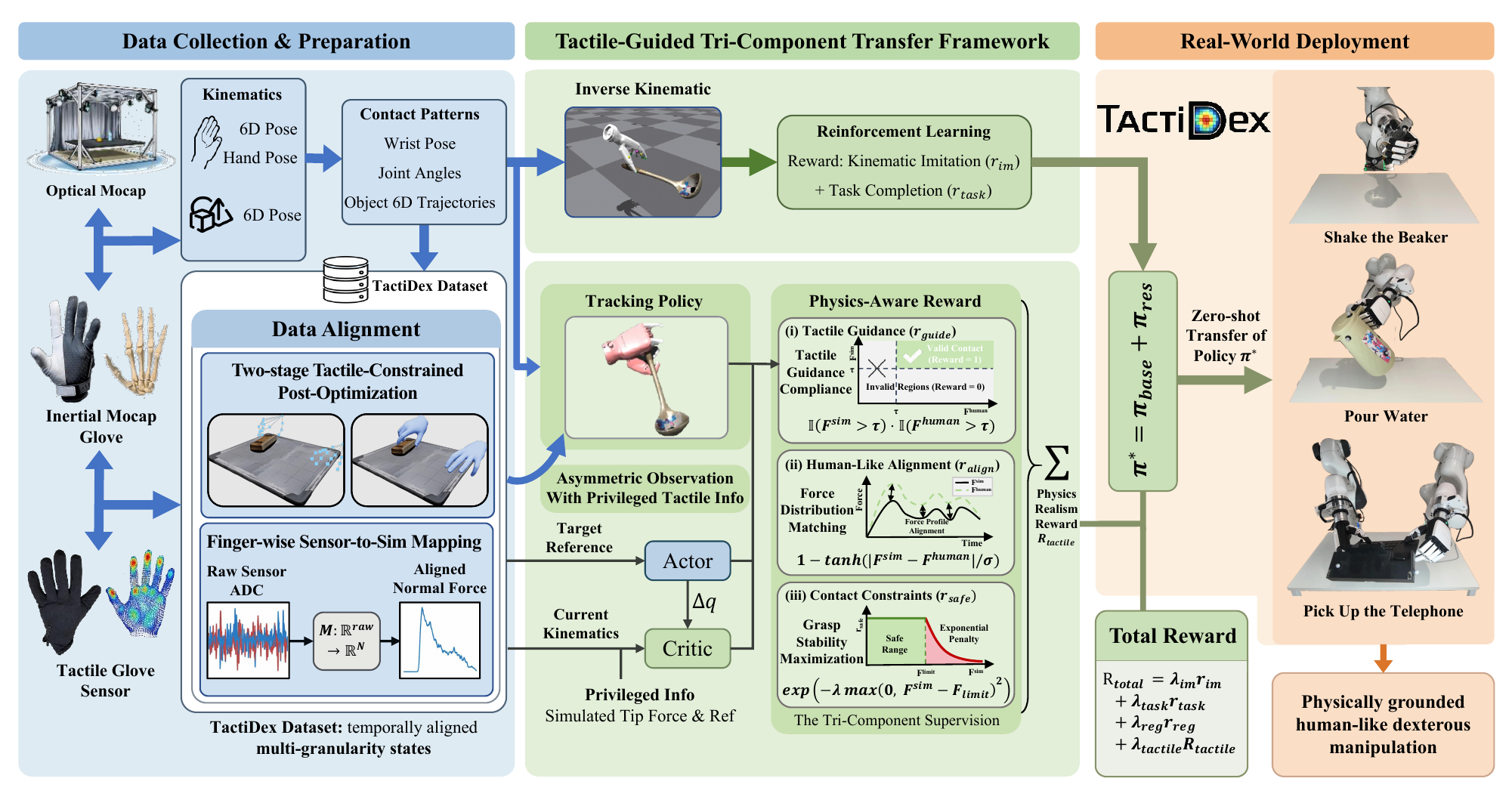}
  \caption{\textbf{Overview of the TactiSkill Pipeline.} 
    \textbf{(Left) Data Collection \& Preparation:} Raw multi-modal signals from human demonstrations undergo a two-stage tactile-constrained post-optimization and sensor-to-sim mapping, yielding the physically aligned TactiDex dataset. 
    \textbf{(Center) Tactile-Guided Tri-Component Transfer Framework:} We train a residual tracking policy using an asymmetric Actor-Critic architecture, where the Critic receives privileged simulated contact forces. A physics-aware tri-component reward (Tactile Guidance, Human-Like Alignment, and Contact Constraints) enforces force-level human-likeness. 
    \textbf{(Right) Real-World Deployment:} The composite policy ($\pi^*$) enables zero-shot sim-to-real transfer, achieving physically grounded, human-like dexterous manipulation on actual robotic hardware across diverse tasks.}
  \label{fig:pipeline}
\end{figure*}

As illustrated in Fig.~\ref{fig:pipeline}, the core objective of our framework is to leverage the TactiDex dataset to train dexterous hands in simulation, enabling them to replicate both the spatiotemporal kinematics and the dynamic contact patterns of human demonstrations. 

\subsection{Overview \& Problem Formulation}
\label{sec:method-overview}

To achieve physically grounded transfer, we model the contact-aware dexterous manipulation as a constrained Markov Decision Process (MDP), denoted by the tuple $\mathcal{M} = \langle \mathcal{S}, \mathcal{A}, \mathcal{P}, \mathcal{R}, \gamma \rangle$. Unlike traditional kinematic-centric setups, we explicitly incorporate high-fidelity tactile signals as Target Contact Features into the state space to enforce physical realism:
\begin{itemize}
    \item \textbf{State Space $\mathcal{S}$:} At timestep $t$, the state $\mathbf{s}_t \in \mathcal{S}$ includes standard proprioception (joint positions $\mathbf{q}_t$, velocities $\dot{\mathbf{q}}_t$) and object states (pose $\mathbf{p}_t$). Crucially, we introduce a Target Tactile Reference denoted as $\mathbf{F}_t^{\text{human}}$ derived from the TactiDex, which guides the robot on the exact normal force each finger should apply at the current moment.
    \item \textbf{Action Space $\mathcal{A}$:} The action $\mathbf{a}_t = \Delta \mathbf{q}_t$ is defined as the residual joint position commands, serving to fine-tune the base kinematic trajectory.
\end{itemize}
The optimization objective is to find an optimal policy $\pi^*(\mathbf{a}_t | \mathbf{s}_t)$ that minimizes kinematic tracking errors while maximizing force consistency with the human demonstration. 

\subsection{Tactile-Guided Tri-Component Human-to-Robot Transfer}
\label{sec:method-framework}

To effectively transfer human demonstrations to morphologically distinct robotic hands, we propose \textbf{TactiSkill}, a novel policy learning method built upon the \textbf{TactiDex} dataset. As depicted in the central block of Fig.~\ref{fig:pipeline}, TactiSkill operates through our proposed \textbf{Tactile-Guided Tri-Component Transfer Framework}.

Building upon recent residual reinforcement learning architectures designed for morphological retargeting \cite{li2025maniptrans}, our final policy $\pi^*$ is composed of a frozen kinematic imitation policy $\pi_{\text{base}}$ and a learnable residual policy $\pi_{\text{res}}$. However, instead of acting merely as a geometric compensator to fix trajectory errors, TactiSkill fundamentally re-purposes $\pi_{\text{res}}$ as a \textit{Force Modulator}. By actively inducing and regulating simulated contact forces, $\pi_{\text{res}}$ bridges the gap between geometric motion mimicry and true physical interaction through a tightly integrated three-step pipeline.

\noindent\textbf{(1) Finger-wise Sensor-to-Sim Data Alignment.}\\
The TactiDex dataset captures raw sensor signals (high-frequency ADC readings), whereas the physics simulation engine computes contact forces in Newtons based on rigid-body dynamics. To establish a unified physical baseline for supervision, we construct a finger-wise, non-linear calibration mapping function $\mathcal{M}: \mathbb{R}^{\text{raw}} \rightarrow \mathbb{R}^N$. This mapping precisely converts the raw human demonstration signals into aligned normal forces consistent with the simulation environment.

\noindent\textbf{(2) Asymmetric Observation with Privileged Tactile Information.}\\
To enhance training stability and sample efficiency, TactiSkill employs an asymmetric Actor-Critic architecture. The \textit{Actor} only receives the target tactile reference $\mathbf{F}_t^{\text{human}}$, mimicking the real-world deployment constraint where the robot possesses a desired force reference but cannot directly foresee the resulting contact dynamics. Conversely, the \textit{Critic} receives \textit{privileged information}—specifically, the real-time simulated fingertip contact force $\mathbf{F}_t^{\text{sim}}$. This enables the Critic to accurately evaluate the value gradient in the latent space, efficiently guiding the Actor to adjust joint poses to approximate the target force distribution.

\noindent\textbf{(3) Physics-Aware Reward: The Tri-Component Supervision.}\\
To enforce force-level human-likeness, the composite tactile reward $R_{\text{tactile}}$ regulates the normal contact force $F_i$ of the $i$-th finger (out of $N_f$ fingers).  It is formulated as a sum of three distinct components, \textit{Tactile Guidance} $r_{\text{guide}}$, \textit{Human-Like Alignment} $r_{\text{align}}$ and \textit{Contact Constraints} $r_{\text{safe}}$ balanced by weights $w_g, w_a, w_s$.  
\begin{equation}
  R_{\text{tactile}} = w_g \cdot r_{\text{guide}} + w_a \cdot r_{\text{align}} + w_s \cdot r_{\text{safe}}.
  \label{eq:tactile_total}
\end{equation}

\textit{i. Tactile Guidance}. Overcomes the common ``air grasping'' issue by providing a binary contact bonus. It acts as a ``contact trigger,'' rewarding the agent only when it successfully replicates an active contact event initiated by the human demonstrator, where $\tau$ is a minimal contact threshold:
\begin{equation}
  r_{\text{guide}} = \frac{1}{N_f} \sum_{i=1}^{N_f} \mathbb{I}\left(F_{i}^{\text{sim}} > \tau\right) \cdot \mathbb{I}\left(F_{i}^{\text{human}} > \tau\right).
  \label{eq:r_guide}
\end{equation}

\textit{ii. Human-Like Alignment}. Merely touching the object is insufficient; the robot must replicate the force modulation patterns. We employ a robust Tanh-based distance metric to minimize the discrepancy between simulated and human forces. Unlike Mean Squared Error, this formulation provides normalized gradients and preserves the force distribution across fingers while resisting sensor outliers ($\sigma$ is a scaling factor):
\begin{equation}
  r_{\text{align}} = \frac{1}{N_f} \sum_{i=1}^{N_f} \left( 1 - \tanh\left(\frac{|F_{i}^{\text{sim}} - F_{i}^{\text{human}}|}{\sigma}\right) \right).
  \label{eq:r_align}
\end{equation}

\textit{iii. Contact Constraints}. Prevents the policy from exploiting the physics engine to generate unrealistically high forces. We define a dynamic safety limit $F_{\text{limit}} = F_{i}^{\text{human}} + \delta$ (where $\delta$ is a safety margin). An exponential penalty is applied only when the simulated force exceeds this limit, ensuring a physically stable and safe Sim-to-Real transfer:
\begin{equation}
  r_{\text{safe}} = \exp\left( - \lambda \sum_{i=1}^{N_f} \max\left(0, F_{i}^{\text{sim}} - F_{\text{limit}}\right)^2 \right).
  \label{eq:r_safe}
\end{equation}

\textbf{Total Reward Formulation.}
\label{sec:method-total-reward}
To ensure the policy converges to a kinematically accurate and physically plausible solution, the final reward function at each timestep $t$ combines our novel tactile supervision with standard kinematic constraints:

\begin{equation}
  R_{total} = \underbrace{\lambda_{im} r_{im} + \lambda_{task} r_{task} + \lambda_{reg} r_{reg}}_{\text{Kinematic \& Task Constraints}} + \underbrace{\lambda_{tactile} R_{tactile}}_{\text{Physical Realism}},
  \label{eq:total_reward}
\end{equation}

While $r_{im}$ (pose tracking), $r_{task}$ (object trajectory), and $r_{reg}$ (energy efficiency) guide the robot to the correct spatiotemporal configuration, the $R_{\text{tactile}}$ term acts as a critical dynamic regularizer, guaranteeing that the interaction forces remain human-like and physically stable.
\section{Experiments}
\label{sec:experiments}

\begin{table*}[tb]
    \caption{Quantitative comparison of TactiSkill against the kinematic baseline and ablation variants over 73 sequences. $\text{SR}_{\text{kin}}$ denotes kinematic success, while $\text{SR}_{\text{tac}}$ enforces strict
  tactile fidelity criteria. $\text{PeakSafe@3N}$ measures the fraction of episodes whose peak force error stays below 3N, and $\text{SafeTac@3N}$ further requires tactile-aware success under the same safety constraint. Best
  results are bolded.}
    \label{tab:ablation}
    \centering
    \resizebox{\textwidth}{!}{
    \begin{tabular}{@{}lcccccccccc@{}}
      \toprule
      Method & \multicolumn{2}{c}{Success Rate} & \multicolumn{2}{c}{Object Tracking} & \multicolumn{2}{c}{Hand Kinematics} & \multicolumn{2}{c}{Tactile Fidelity} & \multicolumn{2}{c}{Safety \& Reliability} \\
      \cmidrule(lr){2-3} \cmidrule(lr){4-5} \cmidrule(lr){6-7} \cmidrule(lr){8-9} \cmidrule(lr){10-11}
       & $\text{SR}_{\text{kin}} \uparrow$ & $\text{SR}_{\text{tac}} \uparrow$ & OTE-t (cm) $\downarrow$ & OTE-r ($^\circ$) $\downarrow$ & MPJPE (cm) $\downarrow$ & TipErr (cm) $\downarrow$ & MTFE (N) $\downarrow$ & Contact
  F1 $\uparrow$ & PeakSafe@3N $\uparrow$ & SafeTac@3N $\uparrow$ \\
      \midrule
      Kinematic Base \cite{li2025maniptrans}
      & 0.7291 & 0.3935 & 1.1947 & 7.4509 & 5.0990 & 5.1833 & 0.1491 & 0.5569 & 0.6366 & 0.3584 \\
      \midrule
      TactiSkill (w/o Bonus)
      & 0.7687 & 0.4193 & 1.2031 & 8.2507 & 5.1743 & 5.1746 & 0.1529 & 0.5425 & 0.5700 & 0.3568 \\

      TactiSkill (w/o Align)
      & 0.7637 & 0.4220 & 1.0828 & 6.9556 & 5.2308 & 5.2643 & 0.1412 & 0.5134 & 0.5988 & 0.3441 \\

      TactiSkill (w/o Safety)
      & 0.7705 & 0.4124 & 1.0265 & 6.3860 & \textbf{5.0327} & 5.2398 & \textbf{0.0921} & 0.5686 & 0.6870 & 0.3872 \\

      \midrule
      \textbf{TactiSkill (Full)}
      & \textbf{0.8195} & \textbf{0.6464} & \textbf{0.9577} & \textbf{6.1974} & 5.0915 & \textbf{5.0793} & 0.1236 & \textbf{0.7384} & \textbf{0.7280} & \textbf{0.5357} \\
      \bottomrule
    \end{tabular}
    }
  \end{table*}

\begin{figure*}[t]
  \centering
  \includegraphics[width=\linewidth]{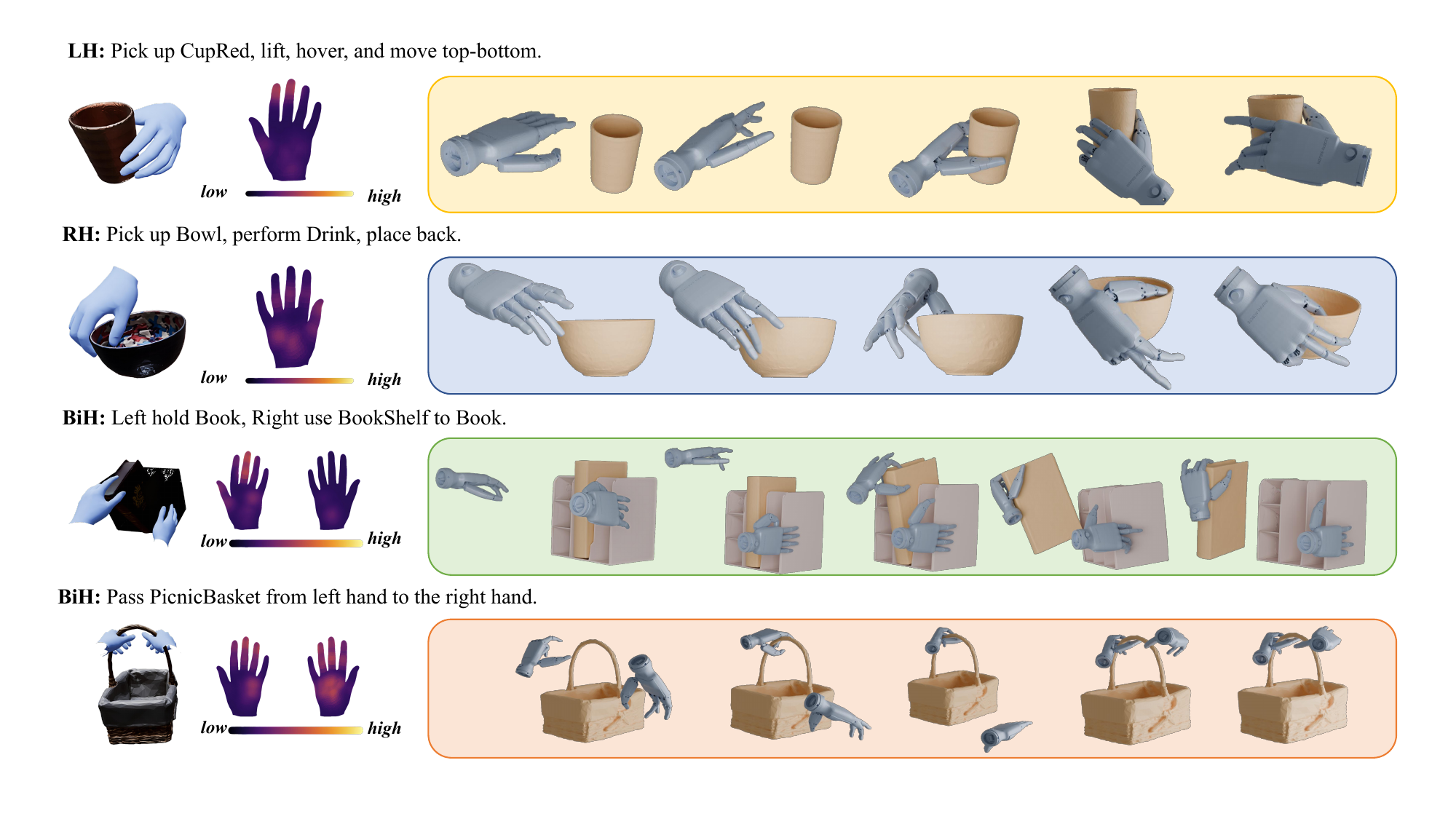}
  \caption{\textbf{Qualitative Results of TactiSkill.} Transfer sequences for single-hand (LH/RH) and bimanual (BiH) tasks. \textbf{Left:} Optimized human HOI reference and whole-hand tactile prior. \textbf{Right (colored panels):} Five-frame execution rollout of the simulated dexterous hand. Guided by our tactile supervision, the policy establishes physically grounded contact, effectively avoiding unnatural hovering or penetrations.}
\label{fig:qualitative}
\end{figure*}

\begin{figure*}[t]
  \centering
  \includegraphics[width=\linewidth]{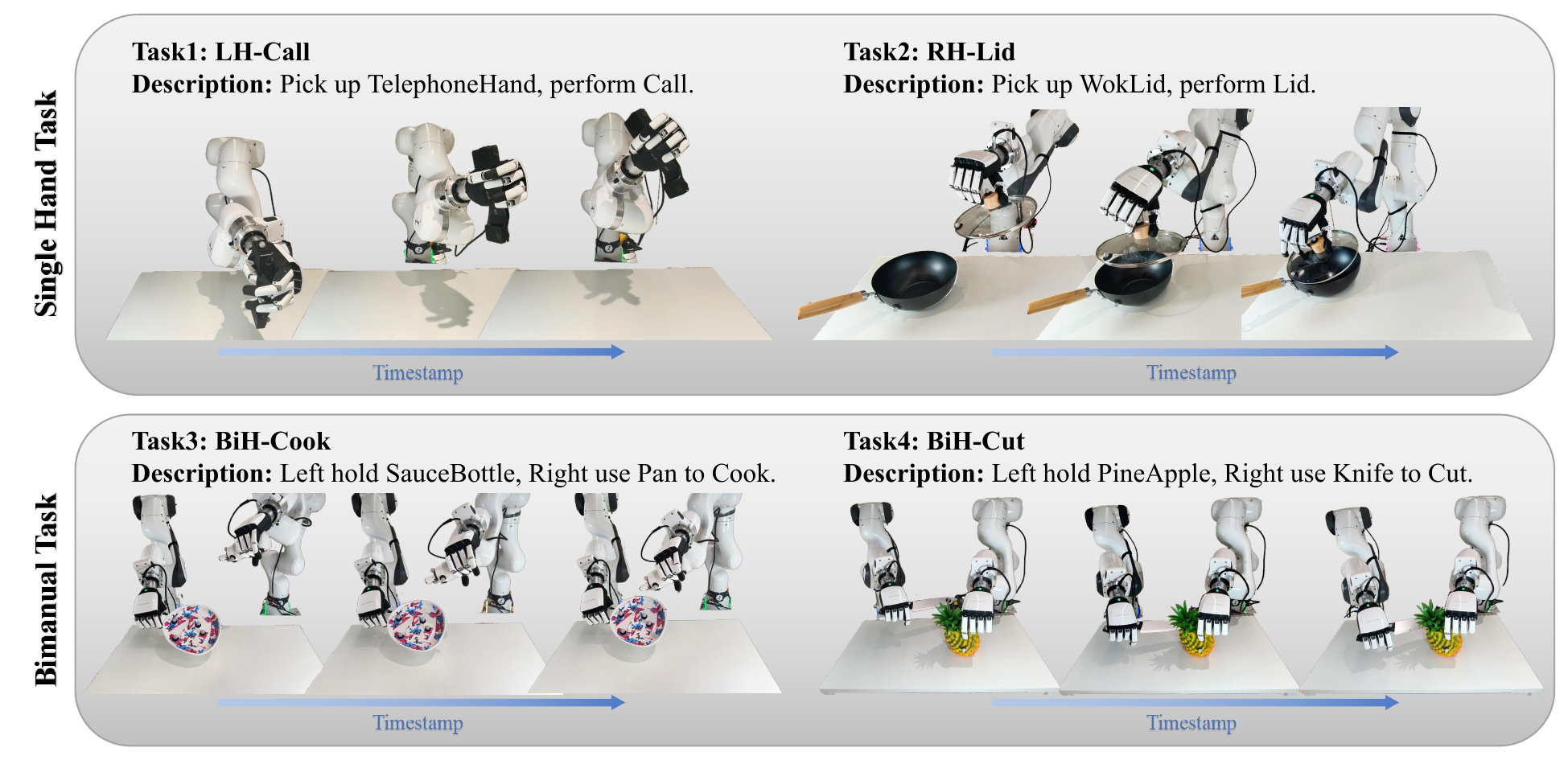}
  \caption{Real-world deployment on our bimanual Franka--Inspire platform. Left: hardware setup with dual Franka arms and Inspire dexterous hands. Right: representative contact-rich execution snapshots.}
  \label{fig:deployment}
\end{figure*}

\subsection{Data Split and Evaluation Metrics}
\label{sec:exp-metrics}
To rigorously evaluate the effectiveness of our tactile-guided transfer framework, we select 73 representative interaction sequences from the \textbf{TactiDex} dataset. This evaluation split encompasses both single-hand and bimanual scenarios, covering diverse object geometries, materials, and varying task complexities. Based on the evaluation dimensions shown in Table~\ref{tab:ablation}, we comprehensively assess the transferred policies across four distinct aspects:

\noindent\textbf{1) Kinematic and Object Tracking.} 
Following established practices \cite{li2025maniptrans}, we evaluate geometric alignment using Object Tracking Error in Translation ($\textbf{OTE}_\textbf{t}$) and Rotation ($\textbf{OTE}_\textbf{r}$). To evaluate how closely the robotic hand replicates human kinematics, we compute the Mean Per-Joint Position Error ($\textbf{MPJPE}$) and specifically the Mean Per-Fingertip Position Error ($\textbf{TipErr}$), as fingertips are the primary medium for interaction.

\noindent\textbf{2) Tactile and Physical Fidelity.} 
To benchmark physical contact-level human-likeness—a core contribution of TactiSkill—we introduce a suite of fine-grained tactile metrics. Let $f_{t,i}^{\text{sim}}$ and $f_{t,i}^{\text{gt}}$ denote the normal force magnitude on the $i$-th finger (out of $N_f$ total fingers) at timestep $t$ (out of $T$ total frames).
\begin{itemize}
    \item \textbf{Mean Tactile Force Error (MTFE).} Computes the mean absolute difference between simulated forces and the aligned human ground truth across all fingers and temporal frames:
    \begin{equation}
        \text{MTFE} = \frac{1}{T \cdot N_f} \sum_{t=1}^{T} \sum_{i=1}^{N_f} \left| f_{t,i}^{\text{sim}} - f_{t,i}^{\text{gt}}. \right|
    \end{equation}
    \item \textbf{Contact F1-Score:} We binarize the continuous force signals into contact states $C_{t,i} = \mathbb{I}(f_{t,i} > \tau_c)$ using a minimal contact threshold $\tau_c$. The F1-Score evaluates the spatial-temporal precision and recall of the simulated contact states against the ground truth, measuring whether the robot engages the correct fingers at the optimal time.
\end{itemize}

\noindent\textbf{3) Safety and Reliability.} 
To explicitly penalize unstable physical spikes and momentary destructive penetrations, we monitor the peak transient force discrepancy $\text{MaxErr} = \max_{t, i} | f_{t,i}^{\text{sim}} - f_{t,i}^{\text{gt}} |$ and introduce two episode-level safety metrics:
\begin{itemize}
    \item \textbf{PeakSafe@3N:} Calculates the fraction of episodes where the maximum instantaneous force error stays strictly below a safety threshold of 3N ($\text{MaxErr} < 3$ N), indicating stable and safe execution without destructive force spikes.
    \item \textbf{SafeTac@3N:} A strictly combined metric requiring the episode to simultaneously achieve tactile-aware success \textit{and} satisfy the peak safety constraint ($\text{MaxErr} < 3$ N).
\end{itemize}

\noindent\textbf{4) Task Success Rate (SR).} 
We evaluate task completion at two distinct levels of strictness.
First, the \textbf{Kinematic Success Rate ($\text{SR}_{\text{kin}}$)} adopts the criteria from \cite{li2025maniptrans}: a trial is successful if $\text{OTE}_\text{r} \le 30^\circ$, $\text{OTE}_\text{t} \le 3$ cm, $\text{MPJPE} \le 8$ cm, and $\text{TipErr} \le 6$ cm. For bimanual tasks, both hands must simultaneously meet these spatial conditions.
Second, to explicitly emphasize physical realism, we employ a stricter \textbf{Tactile-Aware Success Rate ($\text{SR}_{\text{tac}}$)}. A trial achieves tactile success only if it satisfies the kinematic criteria \textit{and} demonstrates high physical fidelity, specifically requiring $\text{MTFE} \le 3.0$ N and $\text{Contact F1} \ge 0.3$.

\subsection{Main Results and Ablative Analysis}
\label{sec:exp-results}
To isolate the contributions of our tactile-guided supervision, we evaluate our \textbf{TactiSkill} framework against a kinematic baseline (\textbf{ManipTrans} \cite{li2025maniptrans}) and conduct an ablation study removing core tactile components: (1) \textbf{w/o Bonus}, (2) \textbf{w/o Align}, and (3) \textbf{w/o Safety} (Table \ref{tab:ablation}).

\noindent\textbf{The Insufficiency of Pure Kinematic Imitation.} 
The purely kinematic baseline (\textbf{Base}) overfits to geometric alignment at the expense of physical realism. While achieving a seemingly adequate $\text{SR}_{\text{kin}}$ (72.91\%) and $\text{OTE}_\text{t}$ (1.1947 cm), this performance is deceptive. By ignoring contact mechanics, the baseline frequently accomplishes goals through ``air grasping'' or mesh penetration. Consequently, its tactile-aware success rate ($\text{SR}_{\text{tac}}$) drops precipitously to 39.35\%, and its safety metrics remain low (Contact F1 = 0.5569, $\text{SafeTac@3N}$ = 35.84\%). This unequivocally indicates that kinematic imitation alone is insufficient for robust tactile manipulation.

\noindent\textbf{The Synergistic Effect of Tactile Guidance.} 
In contrast, the full \textbf{TactiSkill} framework demonstrates that tactile supervision \textit{synergizes} with kinematic tracking to act as a powerful physical regularizer. By enforcing valid contacts, TactiSkill prevents collapsing into invalid geometric optima, attaining the highest $\text{SR}_{\text{kin}}$ (81.95\%) alongside the lowest tracking errors ($\text{OTE}_\text{t}$ = 0.9577 cm, $\text{OTE}_\text{r}$ = $6.1974^\circ$). Most importantly, TactiSkill vastly outperforms all baselines in physical fidelity. It secures the highest Contact F1-score (0.7384) and boosts $\text{SR}_{\text{tac}}$ to 64.64\%. The superior episode-level safety metrics, particularly $\text{PeakSafe@3N}$ (72.80\%) and $\text{SafeTac@3N}$ (53.57\%), validate that our method yields highly reliable and human-like manipulation behavior without destructive force spikes. Specifically, this leap in tactile fidelity highlights a critical shift from passive trajectory following to active force regulation. 

\noindent\textbf{Ablation on Contact Initiation and Alignment.} 
The ablation variants validate our tri-component design. Removing the contact bonus (\textbf{w/o Bonus}) cripples the incentive to initiate stable grasps, causing the Contact F1-score to plummet to 0.5425, and $\text{PeakSafe@3N}$ to drop to 57.00\%. Omitting force alignment (\textbf{w/o Align}) leads to spatially awkward grasps. Failing to match the ground-truth force distribution means the policy cannot utilize anatomically optimal contact patches. This misaligned application severely degrades dynamic stability, reducing Contact F1-score (0.5134) and resulting in a subpar tactile-aware success rate ($\text{SR}_{\text{tac}}$ = 42.20\%).

\noindent\textbf{The Paradox of the Safety Penalty.} 
Interestingly, removing the safety penalty (\textbf{w/o Safety}) yields the lowest mean errors ($\text{MPJPE}$ = 5.0327 cm, $\text{MTFE}$ = 0.0921 N). However, these seemingly superior \textit{average} metrics are highly deceptive. Without an explicit penalty for transient force spikes, the unconstrained policy learns to game the average metric, permitting occasional, highly destructive force impulses. This is evidenced by the degradation in peak safety: compared to the full model, $\text{PeakSafe@3N}$ drops to 68.70\%, and functional safe success ($\text{SafeTac@3N}$) falls significantly to 38.72\%. This highlights our core philosophy: the safety term is crucial not for minimizing isolated local average errors, but for suppressing momentary destructive behaviors, thereby achieving genuinely reliable, safety-aware manipulation at the task level.

\subsection{Qualitative Analysis}
\label{sec:exp-qualitative}
Figure~\ref{fig:qualitative} showcases the qualitative transfer results of our \textbf{TactiSkill} framework evaluated on four representative sequences from the \textbf{TactiDex} dataset. For each single-hand and bimanual task, we present the hand-object interaction (HOI) reference alongside a five-frame rollout of the simulated execution. 

Guided by our tactile-aware supervision, the dexterous hands successfully establish flush and physically stable contact, effectively avoiding unnatural penetrations or hovering. Specifically, in dynamic single-hand tasks (Rows 1-2), our policy seamlessly adapts its multi-fingered posture to diverse object geometries, maintaining secure grasps during complex translations and simulated actions. Furthermore, in demanding bimanual scenarios (Rows 3-4), TactiSkill exhibits exceptional coordination. Whether performing asymmetrical manipulation or dynamic spatial handovers, the framework accurately aligns respective contact patches and regulates interactive forces without dropping the object. Ultimately, this high-fidelity sequential execution underscores the efficacy of our physics-aware tri-component reward in long-horizon environments.

\subsection{Real-World Deployment}
\label{sec:deployment}

We deploy our policy on a bimanual platform with two 7-DoF Franka arms and two Inspire dexterous hands. 
Since the Inspire hand exposes only 6 independent actuators, we retarget the simulated trajectory $\mathbf{q}^{\text{sim}}_{1:T}$ to an actuation trajectory $\tilde{\mathbf{q}}_{1:T}\in\mathbb{R}^6$.

A coarse anchor mapping $\mathbf{q}^{\text{anchor}}_t=\Psi(\mathbf{q}^{\text{sim}}_t)$ provides initialization, and the actuator configuration is expanded to the full kinematic chain via $\Gamma(\cdot)$. 
The retargeting objective is
\begin{equation}
\mathcal{L} =
\mathbb{E}_t
\left[
\sum_f
\|\mathbf{x}^{\text{sim}}_{t,f}-\Phi_f(\Gamma(\tilde{\mathbf{q}}_t))\|_2
+
\lambda_a \|\tilde{\mathbf{q}}_t-\mathbf{q}^{\text{anchor}}_t\|_2^2
+
\lambda_s \|\tilde{\mathbf{q}}_{t+1}-\tilde{\mathbf{q}}_t\|_2^2
\right].
\end{equation}

Arm motions are executed by solving inverse kinematics to track the retargeted wrist pose. 
Tactile sensing is available on the Inspire hands but is not used as feedback in the current deployment.
\balance
\section{Conclusion and Discussion}
\label{sec:conclusion}
\textbf{Conclusion}
In this work, we introduce \textbf{TactiDex}, a tactile-rich benchmark designed to advance dexterous manipulation from purely kinematic imitation toward physically grounded, human-like interaction. TactiDex provides synchronized object trajectories, articulated hand motion, and temporally aligned tactile signals, enabling contact-aware evaluation and learning. Building upon this benchmark, we propose \textbf{TactiSkill}  under the \textbf{Tactile-Guided Tri-Component Transfer Framework}, which explicitly regulates contact formation, force magnitude, and temporal alignment. Through comprehensive quantitative and qualitative experiments, we demonstrate that incorporating structured tactile supervision significantly improves object stability, contact fidelity, and manipulation success over purely kinematic transfer. Overall, our results show that tactile guidance is not merely auxiliary, but essential for bridging the gap between motion imitation and physically realistic dexterous manipulation.

\textbf{Limitations \& Future Work}
Discrepancies in sensor noise, resolution, and physical compliance between simulation and real hardware can degrade the accuracy of learned force interactions when deploying policies directly. To mitigate this, future work will explore on-robot fine-tuning with real tactile feedback, enabling closed-loop adaptation. More broadly, integrating vision, proprioception, and whole-hand tactile sensing within a unified, contact-centric learning framework represents a promising direction toward achieving robust, real-world dexterous manipulation.

\begin{acks}
This work was supported by National Natural Science Foundation of China (62406195, 62303319), HPC Platform of ShanghaiTech University, Core Facility Platform of Computer
Science and Communication of ShanghaiTech University, and Key Laboratory of Intelligent Perception and Human-Machine Collaboration (Shanghaitech University), Ministry of Education, and Shanghai Engineering Research Center of Intelligent Vision and Imaging.
\end{acks}

\clearpage

\bibliographystyle{ACM-Reference-Format}
\bibliography{main}

@String(TOG   = {ACM Trans. Graph.})

@String(TOG   = {ACM TOG})

@inproceedings{wen2024foundationpose,
  title={Foundationpose: Unified 6d pose estimation and tracking of novel objects},
  author={Wen, Bowen and Yang, Wei and Kautz, Jan and Birchfield, Stan},
  booktitle={Proceedings of the IEEE/CVF conference on computer vision and pattern recognition},
  pages={17868--17879},
  year={2024}
}

@article{fang2023anygrasp,
  title={Anygrasp: Robust and efficient grasp perception in spatial and temporal domains},
  author={Fang, Hao-Shu and Wang, Chenxi and Fang, Hongjie and Gou, Minghao and Liu, Jirong and Yan, Hengxu and Liu, Wenhai and Xie, Yichen and Lu, Cewu},
  journal={IEEE Transactions on Robotics},
  volume={39},
  number={5},
  pages={3929--3945},
  year={2023},
  publisher={IEEE}
}

@inproceedings{wen2023bundlesdf,
  title={Bundlesdf: Neural 6-dof tracking and 3d reconstruction of unknown objects},
  author={Wen, Bowen and Tremblay, Jonathan and Blukis, Valts and Tyree, Stephen and M{\"u}ller, Thomas and Evans, Alex and Fox, Dieter and Kautz, Jan and Birchfield, Stan},
  booktitle={Proceedings of the IEEE/CVF Conference on Computer Vision and Pattern Recognition},
  pages={606--617},
  year={2023}
}

@article{yin2025osmo,
  title={OSMO: Open-Source Tactile Glove for Human-to-Robot Skill Transfer},
  author={Yin, Jessica and Qi, Haozhi and Wi, Youngsun and Kundu, Sayantan and Lambeta, Mike and Yang, William and Wang, Changhao and Wu, Tingfan and Malik, Jitendra and Hellebrekers, Tess},
  journal={arXiv preprint arXiv:2512.08920},
  year={2025}
}

@article{lin2023bi,
  title={Bi-touch: Bimanual tactile manipulation with sim-to-real deep reinforcement learning},
  author={Lin, Yijiong and Church, Alex and Yang, Max and Li, Haoran and Lloyd, John and Zhang, Dandan and Lepora, Nathan F},
  journal={IEEE Robotics and Automation Letters},
  volume={8},
  number={9},
  pages={5472--5479},
  year={2023},
  publisher={IEEE}
}

@inproceedings{qi2023hand,
  title={In-hand object rotation via rapid motor adaptation},
  author={Qi, Haozhi and Kumar, Ashish and Calandra, Roberto and Ma, Yi and Malik, Jitendra},
  booktitle={Conference on Robot Learning},
  pages={1722--1732},
  year={2023},
  organization={PMLR}
}

@article{xia2022review,
  title={A review on sensory perception for dexterous robotic manipulation},
  author={Xia, Ziwei and Deng, Zhen and Fang, Bin and Yang, Yiyong and Sun, Fuchun},
  journal={International Journal of Advanced Robotic Systems},
  volume={19},
  number={2},
  pages={17298806221095974},
  year={2022},
  publisher={SAGE Publications Sage UK: London, England}
}

@inproceedings{qin2022dexmv,
  title={Dexmv: Imitation learning for dexterous manipulation from human videos},
  author={Qin, Yuzhe and Wu, Yueh-Hua and Liu, Shaowei and Jiang, Hanwen and Yang, Ruihan and Fu, Yang and Wang, Xiaolong},
  booktitle={European Conference on Computer Vision},
  pages={570--587},
  year={2022},
  organization={Springer}
}

@article{wang2024dexcap,
  title={Dexcap: Scalable and portable mocap data collection system for dexterous manipulation},
  author={Wang, Chen and Shi, Haochen and Wang, Weizhuo and Zhang, Ruohan and Fei-Fei, Li and Liu, C Karen},
  journal={arXiv preprint arXiv:2403.07788},
  year={2024}
}

@inproceedings{li2025maniptrans,
  title={Maniptrans: Efficient dexterous bimanual manipulation transfer via residual learning},
  author={Li, Kailin and Li, Puhao and Liu, Tengyu and Li, Yuyang and Huang, Siyuan},
  booktitle={Proceedings of the IEEE/CVF Conference on Computer Vision and Pattern Recognition},
  pages={6991--7003},
  year={2025}
}

@article{cheng2024open,
  title={Open-television: Teleoperation with immersive active visual feedback},
  author={Cheng, Xuxin and Li, Jialong and Yang, Shiqi and Yang, Ge and Wang, Xiaolong},
  journal={arXiv preprint arXiv:2407.01512},
  year={2024}
}

@inproceedings{bharadhwaj2024roboagent,
  title={Roboagent: Generalization and efficiency in robot manipulation via semantic augmentations and action chunking},
  author={Bharadhwaj, Homanga and Vakil, Jay and Sharma, Mohit and Gupta, Abhinav and Tulsiani, Shubham and Kumar, Vikash},
  booktitle={2024 IEEE International Conference on Robotics and Automation (ICRA)},
  pages={4788--4795},
  year={2024},
  organization={IEEE}
}

@article{qin2023anyteleop,
  title={Anyteleop: A general vision-based dexterous robot arm-hand teleoperation system},
  author={Qin, Yuzhe and Yang, Wei and Huang, Binghao and Van Wyk, Karl and Su, Hao and Wang, Xiaolong and Chao, Yu-Wei and Fox, Dieter},
  journal={arXiv preprint arXiv:2307.04577},
  year={2023}
}

@inproceedings{wang2025dexh2r,
  title={DexH2R: A benchmark for dynamic dexterous grasping in human-to-robot handover},
  author={Wang, Youzhuo and Ye, Jiayi and Xiao, Chuyang and Zhong, Yiming and Tao, Heng and Yu, Hang and Liu, Yumeng and Yu, Jingyi and Ma, Yuexin},
  booktitle={Proceedings of the IEEE/CVF International Conference on Computer Vision},
  pages={12702--12712},
  year={2025}
}

@article{lakshmipathy2025kinematic,
  title={Kinematic motion retargeting for contact-rich anthropomorphic manipulations},
  author={Lakshmipathy, Arjun S and Hodgins, Jessica K and Pollard, Nancy S},
  journal={ACM Transactions on Graphics},
  volume={44},
  number={2},
  pages={1--20},
  year={2025},
  publisher={ACM New York, NY}
}

@article{xin2026analyzing,
  title={Analyzing Key Objectives in Human-to-Robot Retargeting for Dexterous Manipulation},
  author={Xin, Chendong and Yu, Mingrui and Jiang, Yongpeng and Zhang, Zhefeng and Li, Xiang},
  journal={IEEE Robotics and Automation Practice},
  year={2026},
  publisher={IEEE}
}

@inproceedings{xu2023unidexgrasp,
  title={Unidexgrasp: Universal robotic dexterous grasping via learning diverse proposal generation and goal-conditioned policy},
  author={Xu, Yinzhen and Wan, Weikang and Zhang, Jialiang and Liu, Haoran and Shan, Zikang and Shen, Hao and Wang, Ruicheng and Geng, Haoran and Weng, Yijia and Chen, Jiayi and others},
  booktitle={Proceedings of the IEEE/CVF Conference on Computer Vision and Pattern Recognition},
  pages={4737--4746},
  year={2023}
}

@article{wang2022dexgraspnet,
  title={Dexgraspnet: A large-scale robotic dexterous grasp dataset for general objects based on simulation},
  author={Wang, Ruicheng and Zhang, Jialiang and Chen, Jiayi and Xu, Yinzhen and Li, Puhao and Liu, Tengyu and Wang, He},
  journal={arXiv preprint arXiv:2210.02697},
  year={2022}
}

@inproceedings{chao2021dexycb,
  title={Dexycb: A benchmark for capturing hand grasping of objects},
  author={Chao, Yu-Wei and Yang, Wei and Xiang, Yu and Molchanov, Pavlo and Handa, Ankur and Tremblay, Jonathan and Narang, Yashraj S and Van Wyk, Karl and Iqbal, Umar and Birchfield, Stan and others},
  booktitle={Proceedings of the IEEE/CVF conference on computer vision and pattern recognition},
  pages={9044--9053},
  year={2021}
}

@inproceedings{fan2023arctic,
  title={ARCTIC: A dataset for dexterous bimanual hand-object manipulation},
  author={Fan, Zicong and Taheri, Omid and Tzionas, Dimitrios and Kocabas, Muhammed and Kaufmann, Manuel and Black, Michael J and Hilliges, Otmar},
  booktitle={Proceedings of the IEEE/CVF conference on computer vision and pattern recognition},
  pages={12943--12954},
  year={2023}
}

@inproceedings{yang2022oakink,
  title={Oakink: A large-scale knowledge repository for understanding hand-object interaction},
  author={Yang, Lixin and Li, Kailin and Zhan, Xinyu and Wu, Fei and Xu, Anran and Liu, Liu and Lu, Cewu},
  booktitle={Proceedings of the IEEE/CVF conference on computer vision and pattern recognition},
  pages={20953--20962},
  year={2022}
}

@inproceedings{zhan2024oakink2,
  title={Oakink2: A dataset of bimanual hands-object manipulation in complex task completion},
  author={Zhan, Xinyu and Yang, Lixin and Zhao, Yifei and Mao, Kangrui and Xu, Hanlin and Lin, Zenan and Li, Kailin and Lu, Cewu},
  booktitle={Proceedings of the IEEE/CVF Conference on Computer Vision and Pattern Recognition},
  pages={445--456},
  year={2024}
}

@inproceedings{liu2022hoi4d,
  title={Hoi4d: A 4d egocentric dataset for category-level human-object interaction},
  author={Liu, Yunze and Liu, Yun and Jiang, Che and Lyu, Kangbo and Wan, Weikang and Shen, Hao and Liang, Boqiang and Fu, Zhoujie and Wang, He and Yi, Li},
  booktitle={Proceedings of the IEEE/CVF Conference on Computer Vision and Pattern Recognition},
  pages={21013--21022},
  year={2022}
}

@inproceedings{liu2024taco,
  title={Taco: Benchmarking generalizable bimanual tool-action-object understanding},
  author={Liu, Yun and Yang, Haolin and Si, Xu and Liu, Ling and Li, Zipeng and Zhang, Yuxiang and Liu, Yebin and Yi, Li},
  booktitle={Proceedings of the IEEE/CVF Conference on Computer Vision and Pattern Recognition},
  pages={21740--21751},
  year={2024}
}

@inproceedings{liu2025vtdexmanip,
  title={VTDexmanip: A dataset and benchmark for visual-tactile pretraining and dexterous manipulation with reinforcement learning},
  author={Liu, Qingtao and Cui, Yu and Sun, Zhengnan and Li, Gaofeng and Chen, Jiming and Ye, Qi},
  booktitle={The Thirteenth International Conference on Learning Representations},
  year={2025}
}

@article{kim2025tac2motion,
  title={Tac2Motion: Contact-Aware Reinforcement Learning with Tactile Feedback for Robotic Hand Manipulation},
  author={Kim, Yitaek and Rask, Casper Hewson and Sloth, Christoffer},
  journal={arXiv preprint arXiv:2509.17812},
  year={2025}
}

@inproceedings{zhang2023adaptive,
  title={Adaptive barrier smoothing for first-order policy gradient with contact dynamics},
  author={Zhang, Shenao and Jin, Wanxin and Wang, Zhaoran},
  booktitle={International Conference on Machine Learning},
  pages={41219--41243},
  year={2023},
  organization={PMLR}
}

@article{field2025text2touch,
  title={Text2Touch: Tactile In-Hand Manipulation with LLM-Designed Reward Functions},
  author={Field, Harrison and Yang, Max and Lin, Yijiong and Psomopoulou, Efi and Barton, David and Lepora, Nathan F},
  journal={arXiv preprint arXiv:2509.07445},
  year={2025}
}

@inproceedings{bianchini2023simultaneous,
  title={Simultaneous learning of contact and continuous dynamics},
  author={Bianchini, Bibit and Halm, Mathew and Posa, Michael},
  booktitle={Conference on Robot Learning},
  pages={3966--3978},
  year={2023},
  organization={PMLR}
}

@inproceedings{kwon2021h2o,
  title={H2o: Two hands manipulating objects for first person interaction recognition},
  author={Kwon, Taein and Tekin, Bugra and St{\"u}hmer, Jan and Bogo, Federica and Pollefeys, Marc},
  booktitle={Proceedings of the IEEE/CVF international conference on computer vision},
  pages={10138--10148},
  year={2021}
}

@article{tian2024gaze,
  title={Gaze-guided hand-object interaction synthesis: Dataset and method},
  author={Tian, Jie and Ji, Ran and Yang, Lingxiao and Ni, Suting and Ma, Yuexin and Xu, Lan and Yu, Jingyi and Shi, Ye and Wang, Jingya},
  journal={arXiv preprint arXiv:2403.16169},
  year={2024}
}

@article{hoque2025egodex,
  title={Egodex: Learning dexterous manipulation from large-scale egocentric video},
  author={Hoque, Ryan and Huang, Peide and Yoon, David J and Sivapurapu, Mouli and Zhang, Jian},
  journal={arXiv preprint arXiv:2505.11709},
  year={2025}
}

@inproceedings{hasson2019learning,
  title={Learning joint reconstruction of hands and manipulated objects},
  author={Hasson, Yana and Varol, Gul and Tzionas, Dimitrios and Kalevatykh, Igor and Black, Michael J and Laptev, Ivan and Schmid, Cordelia},
  booktitle={Proceedings of the IEEE/CVF conference on computer vision and pattern recognition},
  pages={11807--11816},
  year={2019}
}

@inproceedings{arunachalam2023dexterous,
  title={Dexterous imitation made easy: A learning-based framework for efficient dexterous manipulation},
  author={Arunachalam, Sridhar Pandian and Silwal, Sneha and Evans, Ben and Pinto, Lerrel},
  booktitle={2023 ieee international conference on robotics and automation (icra)},
  pages={5954--5961},
  year={2023},
  organization={IEEE}
}

@article{li2023dexdeform,
  title={Dexdeform: Dexterous deformable object manipulation with human demonstrations and differentiable physics},
  author={Li, Sizhe and Huang, Zhiao and Chen, Tao and Du, Tao and Su, Hao and Tenenbaum, Joshua B and Gan, Chuang},
  journal={arXiv preprint arXiv:2304.03223},
  year={2023}
}

@article{chen2024object,
  title={Object-centric dexterous manipulation from human motion data},
  author={Chen, Yuanpei and Wang, Chen and Yang, Yaodong and Liu, C Karen},
  journal={arXiv preprint arXiv:2411.04005},
  year={2024}
}

@inproceedings{dasari2023learning,
  title={Learning dexterous manipulation from exemplar object trajectories and pre-grasps},
  author={Dasari, Sudeep and Gupta, Abhinav and Kumar, Vikash},
  booktitle={2023 IEEE international conference on robotics and automation (ICRA)},
  pages={3889--3896},
  year={2023},
  organization={IEEE}
}

@inproceedings{liu2024parameterized,
  title={Parameterized quasi-physical simulators for dexterous manipulations transfer},
  author={Liu, Xueyi and Lyu, Kangbo and Zhang, Jieqiong and Du, Tao and Yi, Li},
  booktitle={European Conference on Computer Vision},
  pages={164--182},
  year={2024},
  organization={Springer}
}

@article{makoviychuk2021isaac,
  title={Isaac gym: High performance gpu-based physics simulation for robot learning},
  author={Makoviychuk, Viktor and Wawrzyniak, Lukasz and Guo, Yunrong and Lu, Michelle and Storey, Kier and Macklin, Miles and Hoeller, David and Rudin, Nikita and Allshire, Arthur and Handa, Ankur and others},
  journal={arXiv preprint arXiv:2108.10470},
  year={2021}
}

@article{chen2023bi,
  title={Bi-dexhands: Towards human-level bimanual dexterous manipulation},
  author={Chen, Yuanpei and Geng, Yiran and Zhong, Fangwei and Ji, Jiaming and Jiang, Jiechuang and Lu, Zongqing and Dong, Hao and Yang, Yaodong},
  journal={IEEE Transactions on Pattern Analysis and Machine Intelligence},
  volume={46},
  number={5},
  pages={2804--2818},
  year={2023},
  publisher={IEEE}
}

@inproceedings{ding2021sim,
  title={Sim-to-real transfer for robotic manipulation with tactile sensory},
  author={Ding, Zihan and Tsai, Ya-Yen and Lee, Wang Wei and Huang, Bidan},
  booktitle={2021 IEEE/RSJ International Conference on Intelligent Robots and Systems (IROS)},
  pages={6778--6785},
  year={2021},
  organization={IEEE}
}

@article{miller2025enhancing,
  title={Enhancing Tactile-based Reinforcement Learning for Robotic Control},
  author={Miller, Elle and McInroe, Trevor and Abel, David and Mac Aodha, Oisin and Vijayakumar, Sethu},
  journal={arXiv preprint arXiv:2510.21609},
  year={2025}
}

@article{han2025zero,
  title={Zero-shot Sim2Real Transfer for Magnet-Based Tactile Sensor on Insertion Tasks},
  author={Han, Beining and Joshi, Abhishek and Deng, Jia},
  journal={arXiv preprint arXiv:2505.02915},
  year={2025}
}

@inproceedings{su2024sim2real,
  title={Sim2real manipulation on unknown objects with tactile-based reinforcement learning},
  author={Su, Entong and Jia, Chengzhe and Qin, Yuzhe and Zhou, Wenxuan and Macaluso, Annabella and Huang, Binghao and Wang, Xiaolong},
  booktitle={2024 IEEE International Conference on Robotics and Automation (ICRA)},
  pages={9234--9241},
  year={2024},
  organization={IEEE}
}

@inproceedings{ding2020sim,
  title={Sim-to-real transfer for optical tactile sensing},
  author={Ding, Zihan and Lepora, Nathan F and Johns, Edward},
  booktitle={2020 IEEE International Conference on Robotics and Automation (ICRA)},
  pages={1639--1645},
  year={2020},
  organization={IEEE}
}

@article{tang2025visual,
  title={Visual--Tactile Fusion and SAC-Based Learning for Robot Peg-in-Hole Assembly in Uncertain Environments},
  author={Tang, Jiaxian and Yuan, Xiaogang and Li, Shaodong},
  journal={Machines},
  volume={13},
  number={7},
  pages={605},
  year={2025},
  publisher={MDPI}
}

@article{li2025visuo,
  title={Visuo-tactile feedback policies for terminal assembly facilitated by reinforcement learning},
  author={Li, Yuchao and Jin, Ziqi and Liu, Jin and Ma, Daolin},
  journal={Frontiers in Robotics and AI},
  volume={12},
  pages={1660244},
  year={2025},
  publisher={Frontiers Media SA}
}

@inproceedings{hansen2022visuotactile,
  title={Visuotactile-rl: Learning multimodal manipulation policies with deep reinforcement learning},
  author={Hansen, Johanna and Hogan, Francois and Rivkin, Dmitriy and Meger, David and Jenkin, Michael and Dudek, Gregory},
  booktitle={2022 International Conference on Robotics and Automation (ICRA)},
  pages={8298--8304},
  year={2022},
  organization={IEEE}
}

@article{huang20243d,
  title={3d-vitac: Learning fine-grained manipulation with visuo-tactile sensing},
  author={Huang, Binghao and Wang, Yixuan and Yang, Xinyi and Luo, Yiyue and Li, Yunzhu},
  journal={arXiv preprint arXiv:2410.24091},
  year={2024}
}

@inproceedings{fu2023safe,
  title={Safe self-supervised learning in real of visuo-tactile feedback policies for industrial insertion},
  author={Fu, Letian and Huang, Huang and Berscheid, Lars and Li, Hui and Goldberg, Ken and Chitta, Sachin},
  booktitle={2023 IEEE International Conference on Robotics and Automation (ICRA)},
  pages={10380--10386},
  year={2023},
  organization={IEEE}
}

@article{zhao2025touch,
  title={Touch begins where vision ends: Generalizable policies for contact-rich manipulation},
  author={Zhao, Zifan and Haldar, Siddhant and Cui, Jinda and Pinto, Lerrel and Bhirangi, Raunaq},
  journal={arXiv preprint arXiv:2506.13762},
  year={2025}
}

@article{liang2022visuo,
  title={Visuo-tactile manipulation planning using reinforcement learning with affordance representation},
  author={Liang, Wenyu and Fang, Fen and Acar, Cihan and Toh, Wei Qi and Sun, Ying and Xu, Qianli and Wu, Yan},
  journal={arXiv preprint arXiv:2207.06608},
  year={2022}
}

@article{yu2023mimictouch,
  title={Mimictouch: Leveraging multi-modal human tactile demonstrations for contact-rich manipulation},
  author={Yu, Kelin and Han, Yunhai and Wang, Qixian and Saxena, Vaibhav and Xu, Danfei and Zhao, Ye},
  journal={arXiv preprint arXiv:2310.16917},
  year={2023}
}

@article{hu2025dexterous,
  title={Dexterous in-hand manipulation of slender cylindrical objects through deep reinforcement learning with tactile sensing},
  author={Hu, Wenbin and Huang, Bidan and Lee, Wang Wei and Yang, Sicheng and Zheng, Yu and Li, Zhibin},
  journal={Robotics and Autonomous Systems},
  volume={186},
  pages={104904},
  year={2025},
  publisher={Elsevier}
}

@article{chen2025sam,
  title={Sam 3d: 3dfy anything in images},
  author={Chen, Xingyu and Chu, Fu-Jen and Gleize, Pierre and Liang, Kevin J and Sax, Alexander and Tang, Hao and Wang, Weiyao and Guo, Michelle and Hardin, Thibaut and Li, Xiang and others},
  journal={arXiv preprint arXiv:2511.16624},
  year={2025}
}

@inproceedings{taheri2020grab,
  title={GRAB: A dataset of whole-body human grasping of objects},
  author={Taheri, Omid and Ghorbani, Nima and Black, Michael J and Tzionas, Dimitrios},
  booktitle={European conference on computer vision},
  pages={581--600},
  year={2020},
  organization={Springer}
}

@article{song2025opentouch,
  title={OPENTOUCH: Bringing Full-Hand Touch to Real-World Interaction},
  author={Song, Yuxin Ray and Li, Jinzhou and Fu, Rao and Murphy, Devin and Zhou, Kaichen and Shiv, Rishi and Li, Yaqi and Xiong, Haoyu and Owens, Crystal Elaine and Du, Yilun and others},
  journal={arXiv preprint arXiv:2512.16842},
  year={2025}
}

@article{li2023object,
  title={Object motion guided human motion synthesis},
  author={Li, Jiaman and Wu, Jiajun and Liu, C Karen},
  journal={ACM Transactions on Graphics (TOG)},
  volume={42},
  number={6},
  pages={1--11},
  year={2023},
  publisher={ACM New York, NY, USA}
}

@article{romero2022embodied,
  title={Embodied hands: Modeling and capturing hands and bodies together},
  author={Romero, Javier and Tzionas, Dimitrios and Black, Michael J},
  journal={arXiv preprint arXiv:2201.02610},
  year={2022}
}

@inproceedings{park2019deepsdf,
  title={Deepsdf: Learning continuous signed distance functions for shape representation},
  author={Park, Jeong Joon and Florence, Peter and Straub, Julian and Newcombe, Richard and Lovegrove, Steven},
  booktitle={Proceedings of the IEEE/CVF conference on computer vision and pattern recognition},
  pages={165--174},
  year={2019}
}

\clearpage
\appendix
\section{TactiDex Details}
\label{sec:supp-dataset}

\subsection{Data Scale and Inventory}
The TactiDex dataset features over 700 meticulously recorded interaction sequences, specifically targeting human-to-robot dexterous manipulation. To ensure comprehensive coverage of contact-rich scenarios, we curated a diverse set of 49 everyday objects spanning multiple categories. 

\begin{figure*}[tb!]
  \centering
  \includegraphics[width=\textwidth]{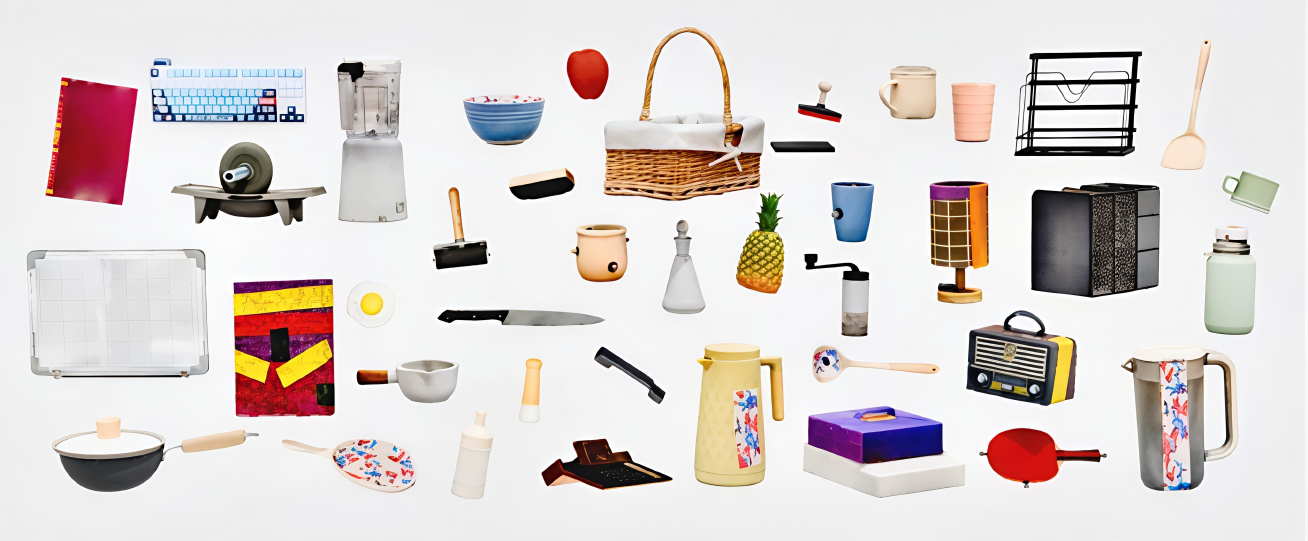}
  \caption{\textbf{Object Inventory of TactiDex.} Rendered meshes of the 49 diverse everyday objects utilized in our data collection. The collection spans a wide range of geometries, physical scales, and functional categories. Notably, our inventory intentionally includes matching multi-part object pairs (e.g., teacups with lids, grinders with rods) to explicitly facilitate complex, long-horizon bimanual interactions and tool use.}
\label{fig:supp_objects}
\end{figure*}

As illustrated in Fig.~\ref{fig:supp_objects}, these objects encompass various geometries, sizes, and physical properties. Rather than being limited to simple rigid bodies, our inventory ranges from delicate items (e.g., eggs) and everyday containers to complex tools and matching multi-part accessories (e.g., a grinder with a rod, or a wok with a lid). This strategic selection explicitly supports the execution of intricate, long-horizon object-object interactions. 

\begin{figure*}[tb]
  \centering
  \includegraphics[width=\textwidth]{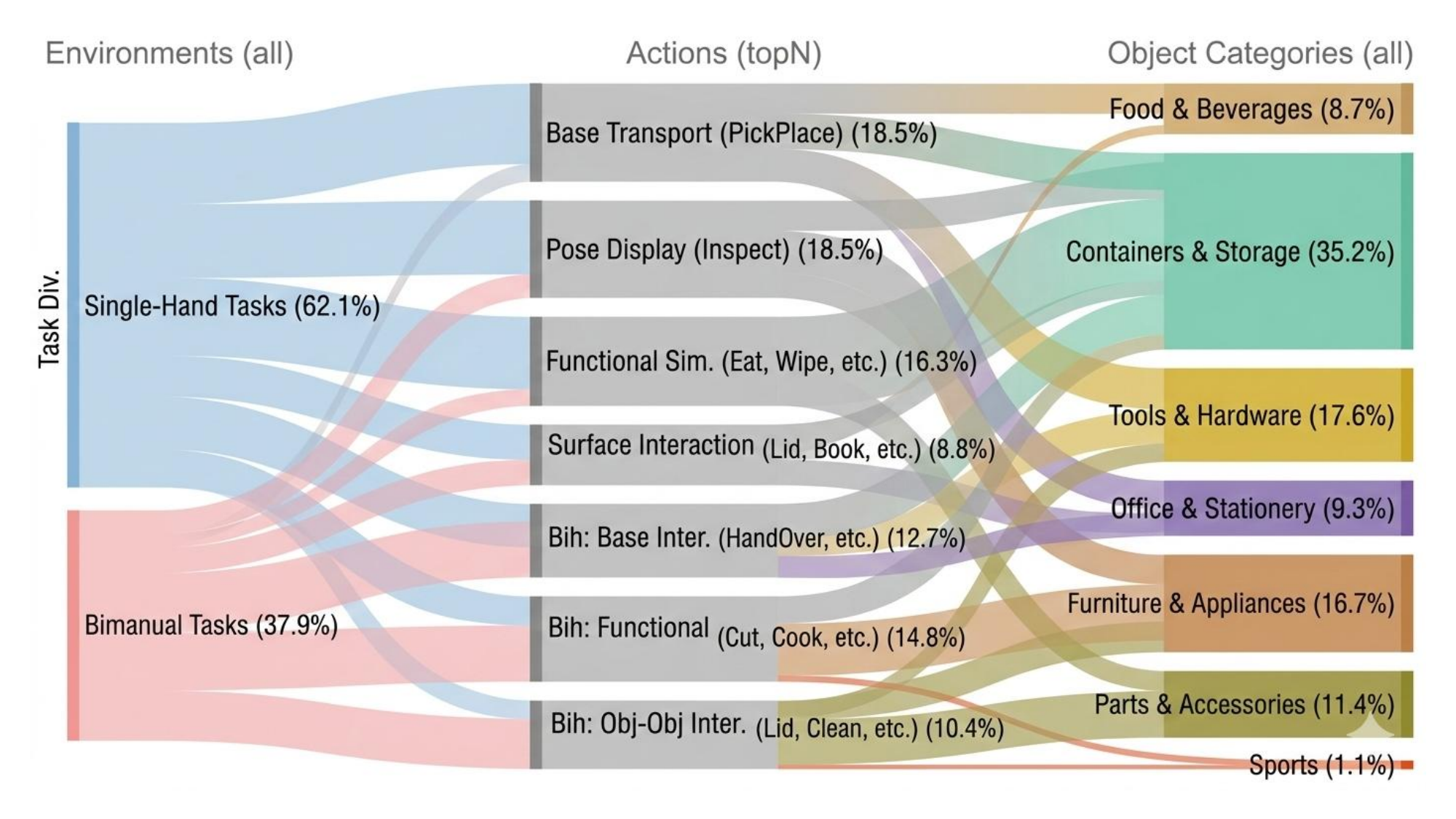}
  \caption{\textbf{Task and Interaction Distribution.} A Sankey diagram illustrating the hierarchical mapping among task actions, semantic grasp types, and object categories across the 700+ sequences in the TactiDex dataset.}
  \label{fig:supp_distribution}
\end{figure*}

Furthermore, to provide a holistic view of the interaction diversity, Fig.~\ref{fig:supp_distribution} visualizes the macroscopic distribution of our recorded sequences. Following previous large-scale interaction datasets, we employ a Sankey diagram to map the explicit hierarchical relationships between task divisions (i.e., single-hand versus bimanual tasks), fine-grained action types (e.g., functional simulation, bimanual handovers, object-object interactions), and specific object categories.

\subsection{Hardware Setup and Capture System}
\label{sec:supp-hardware-details}
To capture high-fidelity, synchronized human demonstrations, we constructed a custom multi-modal capture arena that seamlessly integrates global motion capture, fine-grained hand kinematics, and whole-hand tactile sensing.

\noindent\textbf{Optical Motion Capture System.} 
The global tracking arena is powered by an OptiTrack system equipped with 8 PrimeX 13W (PX13W) cameras, operating at a high sampling rate of 120 fps. Prior to data collection, the capture volume is meticulously calibrated using a specialized vendor-provided active wand. This rigorous calibration process achieves an exceptional mean ray error of exactly 0.2 mm, ensuring sub-millimeter precision for recording both hand trajectories and object 6D poses.

\noindent\textbf{Kinematic and Tactile Dual-Glove Setup.} 
For fine-grained finger kinematics, human demonstrators wear MANUS Metagloves Pro, renowned for their high-precision and low-latency joint tracking. To guarantee anatomical accuracy across different subjects, a standardized calibration routine is strictly performed before every single recording session. 

\begin{figure*}[tb]
  \centering
  \includegraphics[width=0.8\textwidth]{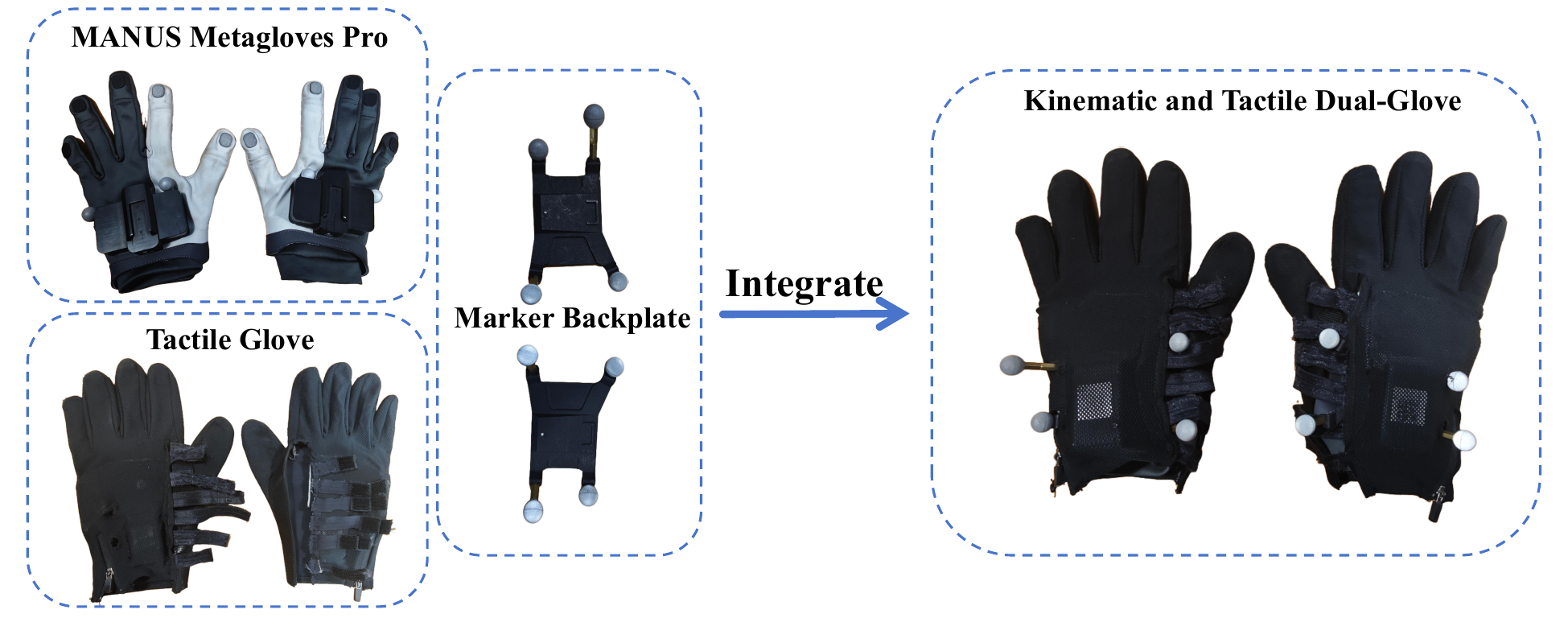}
  \caption{\textbf{Custom Dual-Glove Hardware Integration.} To synchronously capture high-precision kinematics and tactile signals, we developed a customized backplate modification. This design safely routes the tracking markers without interfering with the dense wiring matrix of the outer tactile glove.}
  \label{fig:supp_glove_mod}
\end{figure*}

Simultaneously, the user wears our high-resolution tactile glove on the exterior to capture continuous pressure maps. Integrating these two state-of-the-art systems presented a unique hardware challenge: tracking the Metagloves within the OptiTrack ecosystem requires a rigid marker backplate. To accommodate the outer tactile glove without damaging its delicate internal wiring or compromising sensor fidelity, we engineered a custom, non-destructive mechanical modification for the backplate integration (as illustrated in Fig.~\ref{fig:supp_glove_mod}).

\noindent\textbf{Object Registration and Migration.} 
Beyond capturing raw trajectories, we ensure all object interactions are grounded in a unified physical space. Once the 3D meshes of the 49 objects are generated, we manually align their scale, translation, and rotation within the multi-view OptiTrack coordinate system. This standardized spatial registration process guarantees that the recorded 6D object poses are inherently aligned with the mesh geometry, allowing the entire dataset to be easily migrated and correctly rendered on any new computational platform or simulation environment.

\subsection{Standard Operating Procedure (SOP) and Collection Pipeline}
\label{sec:supp-sop}
Given the unprecedented scale of the TactiDex dataset, relying on manual recording protocols for 700+ multi-modal sequences would be highly susceptible to inconsistencies and human error. 

To systematically standardize our collection process, we designed and deployed a dedicated web-based data collection system. This integrated web platform serves as the central command hub during demonstrations. It dynamically displays the required task instructions to the human subject, tracks the real-time progress of the task inventory, and logs metadata for each recorded trial. By formalizing the entire workflow through this automated SOP interface, we significantly reduced overhead and ensured that task executions remain highly consistent across the entire dataset.

\section{TactiSkill Implementation}
\label{sec:supp-rl}

\subsection{Network Architecture and State-Action Space}
\label{sec:supp-network}

Our TactiSkill framework is implemented using Proximal Policy Optimization (PPO) with an asymmetric Actor-Critic architecture. Both networks are modeled as multi-layer perceptrons (MLPs) featuring hidden dimensions of $[512, 256, 128]$ with ELU activations. 

To bridge the sim-to-real gap, the state space is rigorously decoupled. Taking the 12-DoF Inspire hand as our primary single-hand example, the observation space consists of three distinct components (Note: for bimanual tasks, all dimensions are exactly doubled):

\noindent\textbf{1. Proprioceptive State ($S_{prop} \in \mathbb{R}^{46}$):} Contains the internal kinematic status of the robotic hand.
\begin{itemize}
    \item \textbf{Joint States (36-dim):} Current joint positions $\mathbf{q} \in \mathbb{R}^{12}$, and their trigonometric encodings $\sin(\mathbf{q}), \cos(\mathbf{q}) \in \mathbb{R}^{24}$ to ensure continuity in rotational space.
    \item \textbf{Wrist Base State (10-dim):} The 6D pose (quaternion) and linear/angular velocities of the wrist base, excluding absolute global positioning to promote spatial invariance.
\end{itemize}

\noindent\textbf{2. Target Reference ($S_{target} \in \mathbb{R}^{330}$):} Provides dense spatial and temporal cues from the human demonstration to guide the imitation process.
\begin{itemize}
    \item \textbf{Tactile \& Geometry Prior (133-dim):} A Basis Point Set (BPS) encoding of the object's point cloud ($128$-dim) coupled with the ground-truth target tactile distances for the fingertips ($5$-dim).
    \item \textbf{Future Kinematic Trajectory (197-dim):} The delta kinematics (positions, rotations, linear/angular velocities) over the future horizon for the wrist ($23$-dim), finger joints ($135$-dim), manipulated object ($23$-dim), and the relative object-to-joint transformations ($16$-dim).
\end{itemize}

\noindent\textbf{3. Privileged Information ($S_{priv} \in \mathbb{R}^{49}$):} Accessible exclusively to the Critic during simulation training to accurately estimate the value function.
\begin{itemize}
    \item \textbf{Exact Object Dynamics (14-dim):} Ground-truth Cartesian position ($3$), quaternion ($4$), velocities ($6$), and object weight ($1$).
    \item \textbf{High-Fidelity Contact Force (20-dim):} Simulated fingertip contact forces $\mathbf{F}^{sim} \in \mathbb{R}^{20}$, explicitly capturing both the 3D directional force vectors and their scalar magnitudes across all $5$ fingertips.
    \item \textbf{Internal Physics (15-dim):} Joint velocities $\dot{\mathbf{q}} \in \mathbb{R}^{12}$ and the object's exact Center of Mass (CoM) coordinates ($3$-dim).
\end{itemize}

\noindent\textbf{Network Inputs \& Action Space:}
During training, the Actor network observes only the deployable real-world states: $\mathcal{S}_{actor} = S_{prop} \oplus S_{target} \in \mathbb{R}^{376}$ (or $752$ for bimanual). The Critic network receives the full state including privileged dynamics: $\mathcal{S}_{critic} = S_{prop} \oplus S_{target} \oplus S_{priv} \in \mathbb{R}^{425}$ (or $850$ for bimanual). The action space $\mathcal{A} \in \mathbb{R}^{12}$ outputs the residual target joint positions, which are executed via a low-level PD controller.

\subsection{Hyperparameters and Reward Weights}
\label{sec:supp-hyperparams}

Table~\ref{tab:hyperparameters} comprehensively lists the hyperparameters used for training the residual policy, including the specific weights for our tactile-guided tri-component reward.

\begin{table*}[h]
\caption{Key Hyperparameters and Reward Coefficients for TactiSkill Training}
\label{tab:hyperparameters}
\centering
\begin{tabular}{@{}llc@{}}
\toprule
\textbf{Category} & \textbf{Parameter} & \textbf{Value} \\
\midrule
\multicolumn{3}{l}{\textbf{PPO Algorithm}} \\
 & Learning Rate & $3 \times 10^{-4}$ \\
 & Batch Size (Total) & $32,768$ \\
 & Mini-batch Size & $8,192$ \\
 & Optimization Epochs & $5$ \\
 & Discount Factor ($\gamma$) & $0.99$ \\
 & GAE Parameter ($\lambda$) & $0.95$ \\
\midrule
\multicolumn{3}{l}{\textbf{Tactile Reward}} \\
 & Guidance Weight ($w_g$) & $0.6$ \\
 & Alignment Weight ($w_a$) & $1.4$ \\
 & Safety Penalty Weight ($w_s$) & $0.01$ \\
 & Contact Threshold ($\tau$) & $0.3$ N \\
 & Safety Force Limit ($F_{limit}$) & $40.0$ N \\
\midrule
\multicolumn{3}{l}{\textbf{Kinematic \& Task Reward}} \\
 & Object Tracking Weight (Pos / Rot) & $5.0$ / $3.0$ \\
 & Fingertip Tracking Weight (Thumb / Others) & $0.9$ / $0.6{\sim}0.8$ \\
 & Joint Tracking Weight (Level 1 / Level 2) & $0.5$ / $0.3$ \\
 & Energy Penalty \& Velocity Constraints & $0.05{\sim}0.5$ \\
\bottomrule
\end{tabular}
\end{table*}

The simulation environment is built upon NVIDIA Isaac Gym with a PhysX backend.
\begin{itemize}
    \item \textbf{Physics Parameters:} The simulation operates at $60$ Hz with $2$ sub-steps per frame. To ensure stable grasping and realistic interaction dynamics, we set the lateral friction coefficient of both the fingertips and objects to $1.0$, and the joint damping to $0.5$ Ns/m.
    \item \textbf{Tactile Signal Mapping:} To bridge the gap between human demonstrations and the simulation engine, raw ADC signals from the physical tactile sensors are mapped to simulated normal forces using a linear calibration model: $F_{sim} = k \cdot (ADC_{raw} - ADC_{offset})$, where $k$ is determined via a calibrated force gauge. Within the PhysX engine, this ground-truth force corresponds to the rigid-body contact impulse divided by the simulation time step.
\end{itemize}

\section{Real-World Hardware Deployment Details}
\label{sec:supp-hardware}

While Section 5 in the main text outlines our retargeting objective for zero-shot sim-to-real transfer, this section provides the low-level engineering and communication details required to reproduce our bimanual hardware experiments.

\subsection{Bimanual System Architecture and Arm Control}
Our physical deployment platform consists of two 7-DoF Franka Emika Panda arms and two Inspire dexterous hands. The entire dual-arm control framework is orchestrated using the Deoxys control interface.

To mitigate the sim-to-real global coordinate gap, the Franka arms track relative delta-poses rather than absolute Cartesian coordinates. Before execution, the initial pose of the simulated trajectory is aligned with the physical robot's starting state. At each timestep, the delta position and relative rotation (computed via quaternion transformations) are applied to the physical anchor pose. The arms are then driven by an Operational Space Controller (OSC) operating at a $100$ Hz control loop ($\Delta t = 0.01$ s), which converts Cartesian tracking errors into executable joint torques.

\subsection{Kinematic Optimization for the Inspire Hand}
The simulated hand possesses 12 DoF, whereas the physical Inspire hand is underactuated, driven by only 6 independent linear actuators: Index flexion, Middle flexion, Ring flexion, Pinky flexion, Thumb pitch, and Thumb yaw.

To execute the simulated trajectory $\mathbf{q}^{\text{sim}}_{1:T}$ on the physical hand, we employ an offline optimization-based retargeting pipeline. We utilize \textit{PyTorch Kinematics} to construct the differentiable forward kinematics chain of the Inspire hand. For each trajectory, we initialize a 6-DoF command sequence $\tilde{\mathbf{q}}_{1:T}$ using a heuristic anchor mapping. We then utilize the Adam optimizer (learning rate $= 0.01$, $100$ iterations) to minimize the following composite loss:
\begin{equation}
\mathcal{L} = W_{\text{pos}} \mathcal{L}_{\text{pos}} + W_{\text{anchor}} \mathcal{L}_{\text{anchor}} + W_{\text{smooth}} \mathcal{L}_{\text{smooth}}
\end{equation}
Here, $\mathcal{L}_{\text{pos}}$ penalizes the Cartesian distance between the 12-DoF simulated fingertips and the 6-DoF retargeted fingertips. $\mathcal{L}_{\text{anchor}}$ prevents the solution from drifting too far from the anchor prior, and $\mathcal{L}_{\text{smooth}}$ penalizes temporal jitter. Based on empirical tuning, the weights are set to $W_{\text{pos}} = 50.0$, $W_{\text{anchor}} = 20.0$, and $W_{\text{smooth}} = 10.0$.

\subsection{Command Mapping and Open-Loop Execution}
Once optimized, the 6-DoF joint radians are calibrated and linearly mapped to discrete actuation commands $\in [0, 1000]$. These commands are continuously dispatched to drive the Inspire hand via a $115200$ baud serial interface. 

Crucially, while the physical Inspire hands are equipped with tactile sensors, our current deployment relies exclusively on open-loop execution (i.e., the tactile signals are not fed back into the policy online). The successful sim-to-real transfer demonstrates a core advantage of our TactiSkill framework: by actively regulating simulated contact forces via the tri-component reward during training, the policy intrinsically learns robust, physics-aware force distributions. Consequently, the resulting zero-shot kinematic trajectories naturally establish stable, physically grounded grasps on the real hardware, successfully overcoming typical geometric domain gaps without requiring fragile online tactile feedback loops.

\begin{figure*}[h]
  \centering
  \includegraphics[width=\textwidth]{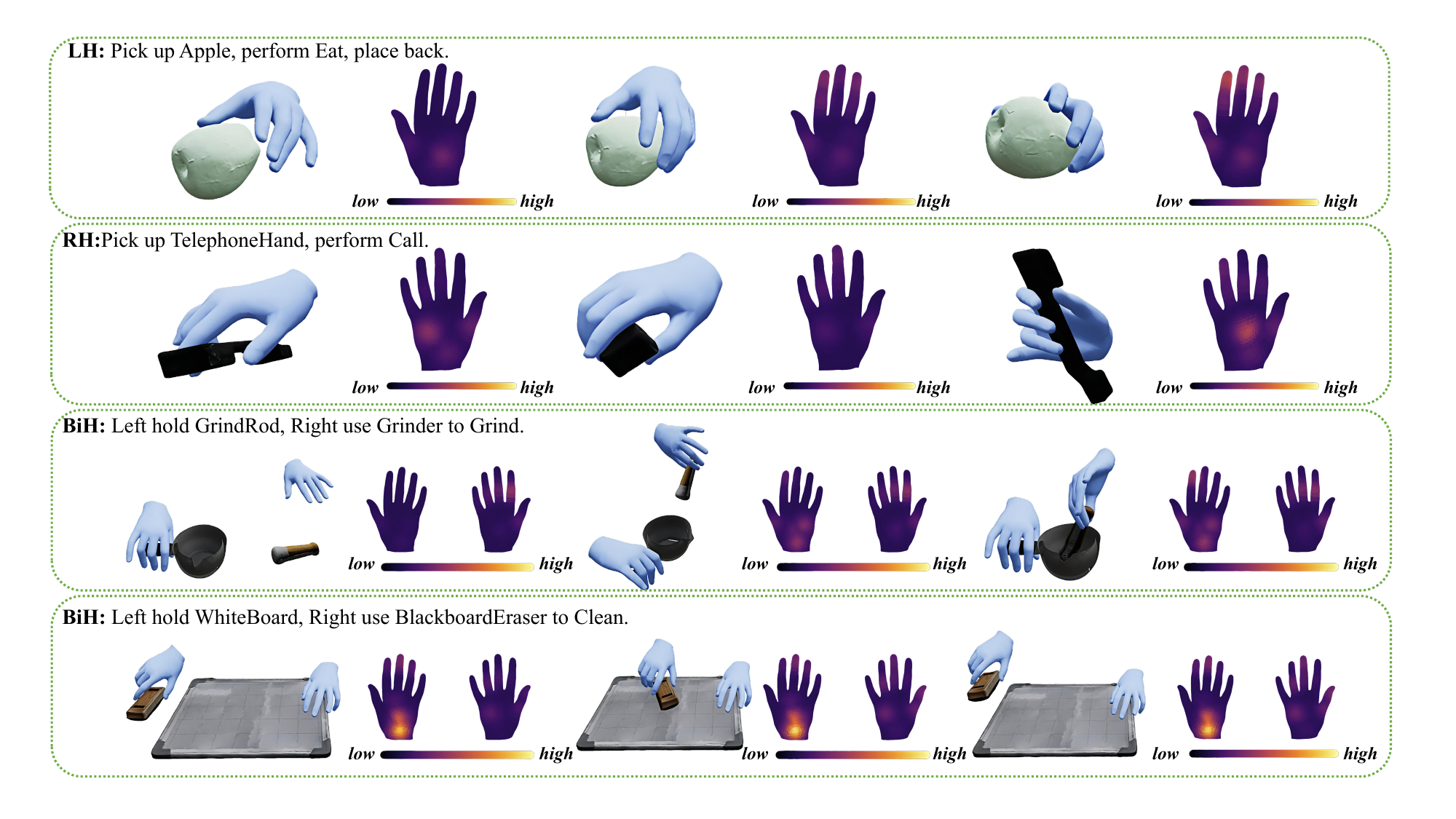} 
  \caption{\textbf{Multi-modal Human Demonstrations from the TactiDex Dataset.} Representative sequences for LH (Apple), RH (TelephoneHand), and BiH (Grinder, WhiteBoard) tasks. Each frame pairs the 3D kinematic hand-object interaction with its synchronized tactile pressure heat map (where purple indicates low pressure and yellow indicates high pressure). These sequences demonstrate how our dataset captures the precise, task-specific force distributions required for complex object manipulation.}
  \label{fig:raw_HOI}
\end{figure*}

\begin{figure*}[h]
  \centering
  \includegraphics[width=\textwidth]{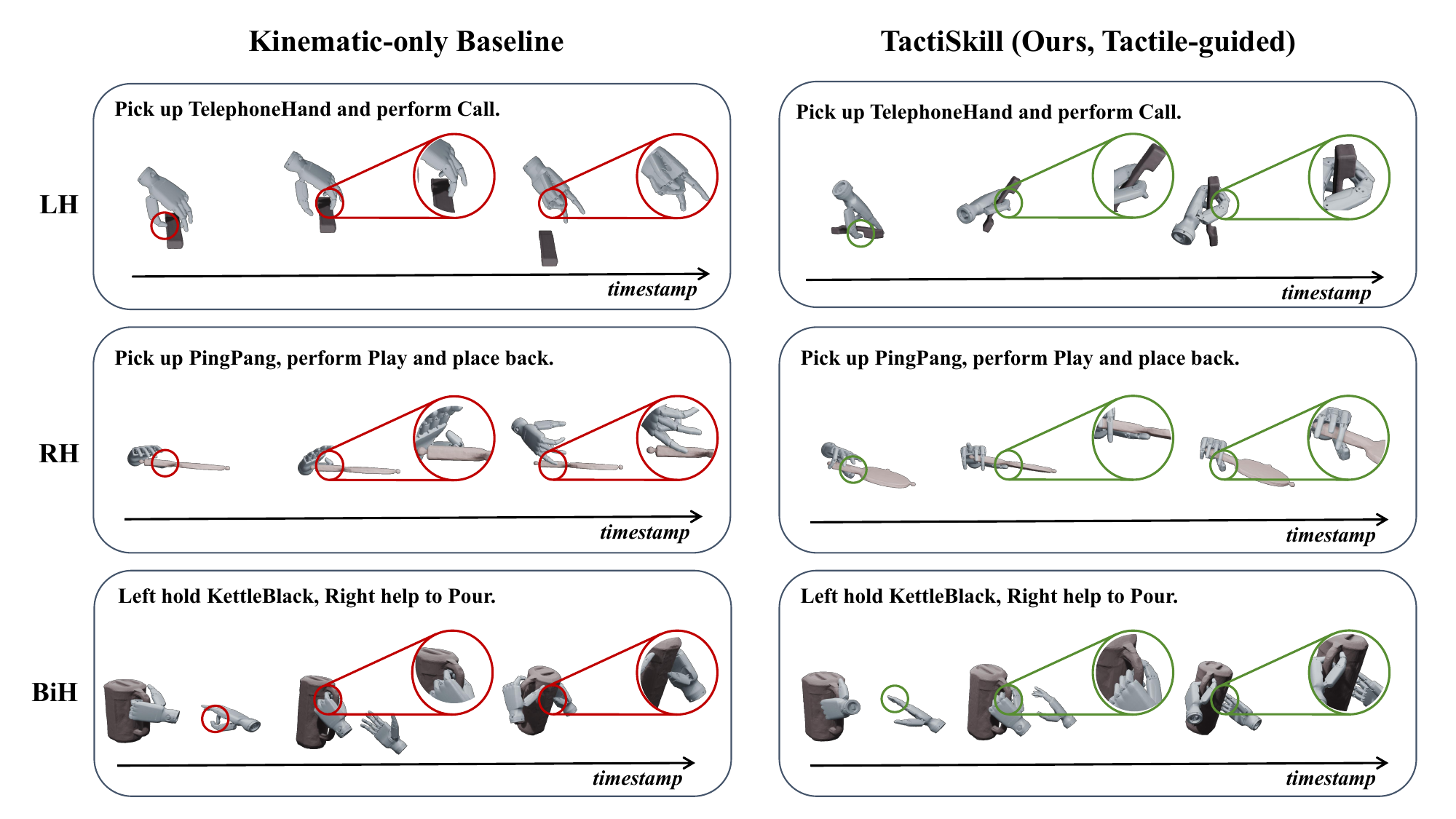} 
  \caption{\textbf{Qualitative Comparison of Contact Quality.} Frame-by-frame rollouts of LH (TelephoneHand), RH (PingPang), and BiH (KettleBlack) tasks. (Left) The kinematic-only baseline often results in unnatural hovering or severe mesh penetration, as explicitly highlighted by the red magnified views. (Right) Our TactiSkill policy seamlessly establishes flush, stable contact with diverse object geometries, effectively preventing physical artifacts as shown in the green magnified views.}
  \label{fig:supp_compare}
\end{figure*}

\section{Extended Qualitative Results and Video Guide}
\label{sec:supp-extra}

\subsection{Visualization of the TactiDex Dataset}
Before evaluating the learned policies, we first visualize the rich, multi-modal human demonstrations captured in our TactiDex dataset. As the foundation of our tactile-guided reinforcement learning framework, the dataset goes beyond standard kinematic tracking by synchronously recording high-resolution contact forces.

As illustrated in Fig.~\ref{fig:raw_HOI}, we present representative sequences across diverse manipulation scenarios, including left-hand (LH), right-hand (RH), and long-horizon bimanual (BiH) tasks. Alongside the 3D kinematic hand-object spatial relationships, we visualize the corresponding human contact heat maps at each timestep (mapped to a purple-to-yellow color scale). These dynamic pressure distributions explicitly reveal the underlying physical intent of the demonstrator. For instance, the heat maps clearly distinguish between the localized fingertip forces required to pinch an apple or hold a telephone, versus the broader palm-level pressure required to firmly press a blackboard eraser. By capturing these nuanced, physics-grounded interactions, the TactiDex dataset provides the essential dense tactile priors that our TactiSkill framework leverages during policy training.

\subsection{Qualitative Comparison of Interaction Quality}
To visually evaluate the impact of our proposed tri-component tactile reward, we provide a localized, frame-by-frame comparison between our \textit{TactiSkill} policy and a standard kinematic-only imitation baseline. As illustrated in Fig.~\ref{fig:supp_compare}, we examine representative left-hand (LH), right-hand (RH), and bimanual (BiH) tasks along their execution timestamps.

Without the explicit guidance of tactile priors, the baseline policy relies entirely on minimizing joint-space discrepancies. This purely kinematic approach frequently suffers from critical physical inconsistencies during complex interactions, as highlighted in the magnified red regions:
\begin{itemize}
    \item \textbf{Kinematic-only Baseline:} The policy fails to respect physical contact boundaries. This results in obvious \textit{unnatural hovering} (e.g., fingers leaving large gaps around the telephone handle or paddle) or \textit{severe mesh penetration} (e.g., fingers aggressively clipping through the kettle handle to artificially secure a grasp).
    \item \textbf{TactiSkill (Ours):} By actively regularizing simulated contact forces during the RL process, our policy ensures the robotic fingertips establish \textit{flush, physically grounded contacts} (highlighted in the magnified green regions). Our tactile-guided policy gracefully conforms to the object's surface boundaries, maintaining stable and realistic grasps across all manipulation stages.
\end{itemize}

\subsection{Supplementary Video Guide}
The dynamic performance, temporal stability, and zero-shot real-world deployment of our framework are best appreciated in motion. We strongly encourage the reviewers to refer to the accompanying \textbf{Supplementary Video} for a comprehensive evaluation of the uncurated policy rollouts and hardware experiments.

\end{document}